\pdfoutput=1

\documentclass[11pt]{article}

\usepackage[preprint]{acl}

\usepackage{times}
\usepackage{latexsym}
\usepackage{enumitem}

\usepackage[T1]{fontenc}

\usepackage[utf8]{inputenc}
\usepackage{booktabs}

\usepackage{microtype}

\usepackage{inconsolata}

\usepackage{graphicx}
\usepackage{multirow}

\usepackage{todonotes}

\title{A Generative-AI-Driven Claim Retrieval System Capable of Detecting and Retrieving Claims from Social Media Platforms in Multiple Languages}

\author{Ivan Vykopal$^{1, 2}$, Martin Hyben$^{2}$, Robert Moro$^{2}$, Michal Gregor$^{2}$ \and \textbf{Jakub Simko}$^{2}$ \\
    $^{1}$ Faculty of Information Technology, Brno University of Technology, Brno, Czech Republic\\
    $^{2}$ Kempelen Institute of Intelligent Technologies, Bratislava, Slovakia \\
    \texttt{\{name.surname\}@kinit.sk}
  }

\begin{document}
\maketitle
\begin{abstract}

Online disinformation poses a global challenge, placing significant demands on fact-checkers who must verify claims efficiently to prevent the spread of false information. A major issue in this process is the redundant verification of already fact-checked claims, which increases workload and delays responses to newly emerging claims. This research introduces an approach that retrieves previously fact-checked claims, evaluates their relevance to a given input, and provides supplementary information to support fact-checkers. Our method employs large language models (LLMs) to filter irrelevant fact-checks and generate concise summaries and explanations, enabling fact-checkers to faster assess whether a claim has been verified before. In addition, we evaluate our approach through both automatic and human assessments, where humans interact with the developed tool to review its effectiveness. Our results demonstrate that LLMs are able to filter out many irrelevant fact-checks and, therefore, reduce effort and streamline the fact-checking process.

\end{abstract}

\section{Introduction}

\begin{figure}[t!]
    \vspace{-5mm}
    \hspace{-3mm}
    \includegraphics[width=1.05\columnwidth]{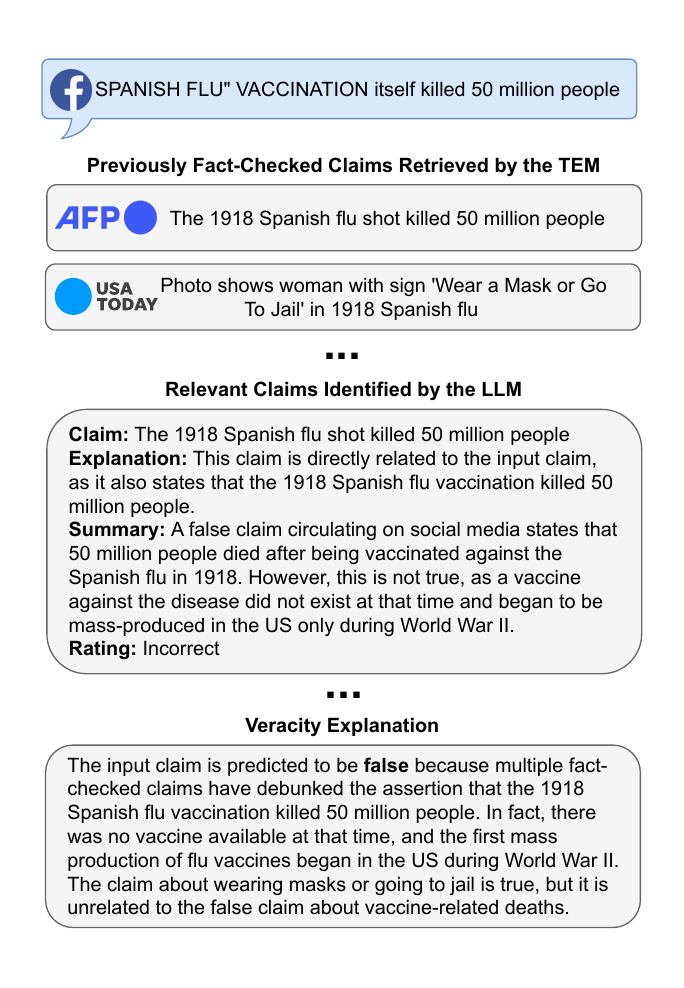}
    \vspace{-8mm}
    \caption{An example of a post with two fact-checked claims retrieved by the embedding model. The LLM selects the relevant claim, explains its choice, summarizes the fact-check article, and predicts the post's veracity.}
    \label{fig:example}
\end{figure}

The rise of social media has accelerated the spread of false information, posing significant societal, economic and public health risks~\cite{10.1145/3161603}. This challenge is further compounded by the multilingual nature of false information, making fact-checking a complex and resource-intensive task for fact-checkers. Fact-checkers often struggle to verify claims across languages, particularly in low-resource settings where limited fact-checking support exists~\cite{hrckova2024autonomationautomationactivitiesneeds}. To address this issue, it is crucial to develop multilingual fact-checking approaches that can assist fact-checkers to identify and verify misinformation efficiently.

One of the key tasks in fact-checking is claim retrieval, also known as previously fact-checked claim retrieval~\cite{pikuliak-etal-2023-multilingual}, where the goal is to identify fact-checks from a database that are the most similar to a given input. This task is crucial, as many claims are not entirely new but rather rephrased or repeated versions of previously debunked misinformation. Efficient retrieval enables fact-checkers to quickly detect repeated claims, reduce redundant efforts, and prioritize emerging or complex claims~\cite{hrckova2024autonomationautomationactivitiesneeds}. However, retrieved results may include fact-checks that are only loosely related or irrelevant, increasing the workload. To mitigate this, LLMs can be leveraged to assess the relevance of retrieved fact-checks, thereby streamlining the review process~\cite{vykopal2025largelanguagemodelsmultilingual}.

In this paper, we propose a novel pipeline for retrieving previously fact-checked claims and assisting fact-checkers in assessing their relevance to a given query. Our experiments cover more than 10 languages from diverse linguistic families and scripts, including low- and high-resource languages. We analyze the ability of language models to retrieve relevant fact-checks while incorporating summarization and explanation. An example is shown in Figure~\ref{fig:example}. In addition, we evaluate LLMs' performance to determine the veracity based on retrieved fact-checks and supporting information.\footnote{The data are available at Zenodo upon request \textit{for research purposes only}: \url{https://zenodo.org/records/15267292}. The source code is available at: \url{https://github.com/kinit-sk/claim-retrieval}.}

Our contributions are as follows:

\begin{itemize}
    \item We provide a novel \textbf{\textit{AFP-Sum}} dataset consisting of around 19K fact-checking articles along with their summaries across 23 languages. Additionally, we created a subset of 2300 fact-checks in 23 languages along with the summaries in the original language and translated summaries in English.
    \item We evaluated multiple text embedding models (TEMs) for retrieving previously fact-checked claims across 20 languages and the capabilities of TEMs for filtering fact-checks based on instructions in the natural language.
    \item We proposed a novel pipeline for incorporating LLMs into the verification process by employing LLMs for identifying relevant previously fact-checked claims, providing fact-check summaries and predicting the veracity of a given claim based on the previously retrieved fact-checks. 
\end{itemize}

\section{Related Work}

\paragraph{Previously Fact-Checked Claim Retrieval.}

Previously fact-checked claim retrieval, also known as verified claim retrieval~\cite{10.1007/978-3-030-58219-7} or claim-matching~\cite{kazemi-etal-2021-claim}, aims to reduce fact-checkers' workload by retrieving similar, already verified claims. While most research focused on monolingual settings~\cite{shaar2020knownliedetectingpreviously, shaar2022assistinghumanfactcheckersdetecting, hardalov2022crowdcheckeddetectingpreviouslyfactchecked}, multilingual retrieval remains underexplored~\cite{vykopal2024generativelargelanguagemodels}. Recent work, such as \citet{pikuliak-etal-2023-multilingual}, introduces the \textbf{\textit{MultiClaim}} dataset for multilingual claim retrieval, evaluating various TEMs for ranking fact-checked claims in monolingual and cross-lingual contexts.

Recent advancements in LLMs present new opportunities for enhancing claim retrieval. Existing approaches primarily rely on two strategies. The first involves \textit{textual entailment}, where models classify the entailment between an input claim and a fact-check into three categories~\cite{10.1145/3589335.3651910, 10.1145/3589335.3651504}. In contrast, the second strategy employs \textit{generative re-ranking} to rank the previously fact-checked claims based on the conditional probabilities generated by LLMs, which are used to prioritize more relevant claims~\cite{shliselberg2022riet, neumann-etal-2023-deep}.

\paragraph{Fact-Checking Pipelines \& Tools.}

With the growing importance of online fact-checking, numerous pipelines have been developed to combat misinformation. Many of these systems rely on retrieving the evidence based on a given claim and leveraging LLMs to asses veracity and provide justifications. However, most research has predominantly focused on English~\cite{10.14778/3137765.3137815, 10.1145/3292500.3330935, li-etal-2024-self} or Arabic~\cite{jaradat-etal-2018-claimrank, althabiti2024takeedgenerativefactcheckingarabic, 10.1145/3539618.3591815}, with fewer studies dedicated to other languages.

Several online tools have been developed to address false information. WeVerify\footnote{\url{https://weverify.eu/}} provides a suite of tools for identifying false information, including image analysis for detecting manipulated content. In addition, BRENDA~\cite{10.1145/3397271.3401396} assesses the credibility of claims, helping users evaluate online information. Furthermore, the models trained by \citet{pikuliak-etal-2023-multilingual} for retrieving previously fact-checked claims have been integrated into the Fact-Check Finder\footnote{\url{https://fact-check-finder.kinit.sk/}}, a tool designed to assist fact-checkers in identifying relevant fact-checked claims across multiple languages.



\paragraph{Multilingual Summarization.}

Multilingual summarization has been propelled by the development of extensive datasets and the application of LLMs~\cite{scialom-etal-2020-mlsum, hasan-etal-2021-xl, bhattacharjee-etal-2023-crosssum}. These resources have enabled the fine-tuning of multilingual models like mT5~\cite{xue-etal-2021-mt5}, which demonstrate competitive performance in both multilingual and low-resource summarization tasks. Furthermore, studies have explored the zero-shot and few-shot capabilities of LLMs such as GPT-3.5 and GPT-4 in cross-lingual summarization, highlighting their potential to handle diverse language pairs without extensive fine-tuning~\cite{wang-etal-2023-zero}. Efforts to enhance factual consistency in multilingual summarization have also been made, exemplified by the use of multilingual models to improve the reliability of machine-generated summaries across languages~\cite{aharoni-etal-2023-multilingual}.

\section{Methodology}
\label{sec:methodology}

Our experiments aim to evaluate the capabilities of TEMs and LLMs in assisting fact-checkers by providing additional information. This includes retrieving the most similar previously fact-checked claims, summarizing fact-checking articles along with their ratings, and potentially predicting the veracity of a given input based on the retrieved information. Much of this process can be automated using LLMs, thereby reducing the effort required from fact-checkers to identify relevant fact-checks. 

\begin{figure*}
    \centering
    \includegraphics[width=0.75\textwidth]{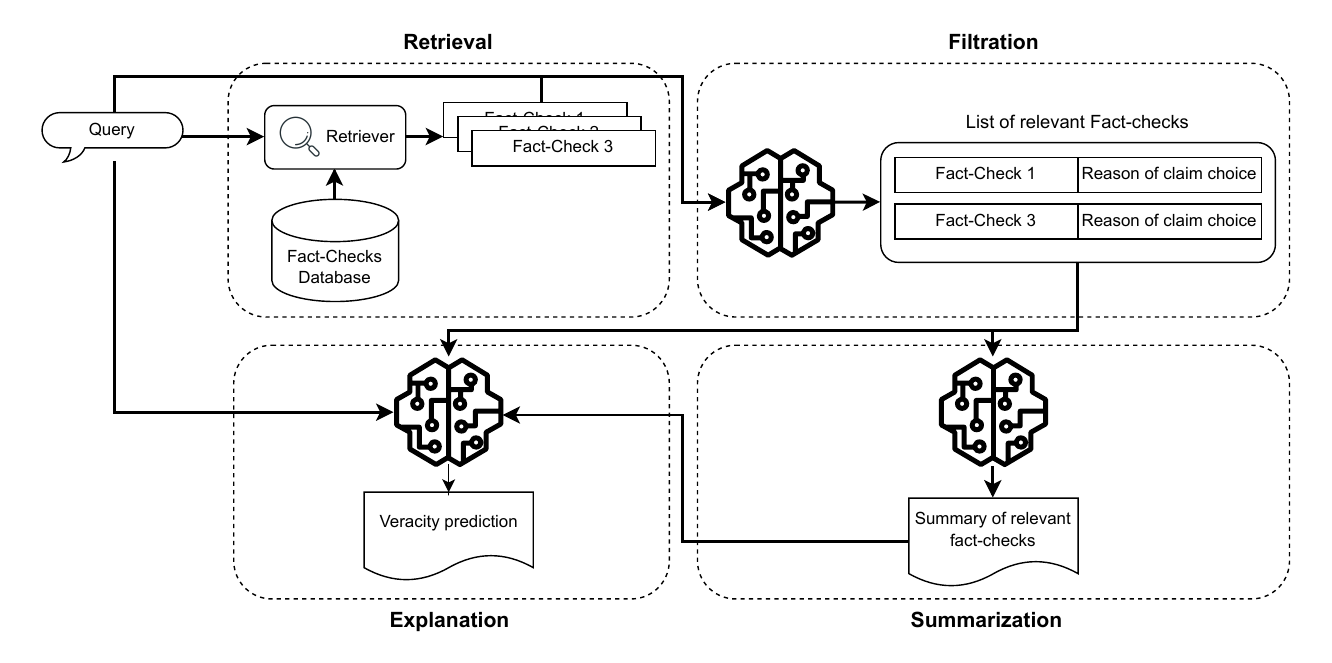}
    \vspace{-3mm}
    \caption{Our proposed pipeline consisting of (1) retrieval of the top N most similar fact-checks, (2) identifying relevant fact-checked claims, (3) summarizing relevant fact-checking articles, and (4) predicting the veracity of the query along with the explanation.}
    \label{fig:uc1-pipeline}
\end{figure*}

Our proposed pipeline, illustrated in Figure~\ref{fig:uc1-pipeline}, consists of four key steps: \textit{retrieval} (Section~\ref{sec:retrieval}), \textit{filtration} (Section~\ref{sec:filtration}), \textit{summarization} (Section~\ref{sec:summarization}) and \textit{veracity prediction} (Section~\ref{sec:veracity}). In the \textit{retrieval} step, the TEM retrieves the top $K$ most similar fact-checks based on a given input. The \textit{filtration} step then refines these results by using the LLM to identify only the fact-checks that are directly relevant, providing explanations for its selection and filtering out irrelevant claims. In the \textit{summarization} step, the LLM generates concise summaries of the relevant fact-checking articles. Finally, in the \textit{veracity prediction} step, the LLM leverages the retrieved fact-checks, their ratings, and the generated summaries to assess and predict the veracity of the given input based on the available information.

In addition, we provide an overview of the datasets (Section~\ref{sec:dataset}) and models (Section~\ref{sec:models}) used in our experiments. We also detailed the evaluation for each step of the pipeline in Section~\ref{sec:eval}.

\subsection{Datasets}
\label{sec:dataset}

\paragraph{MultiClaim.}

We selected the \textbf{\textit{MultiClaim}} dataset~\cite{pikuliak-etal-2023-multilingual} to evaluate the efficiency of embedding models and LLMs in retrieving previously fact-checked claims and assessing claim veracity. \textbf{\textit{MultiClaim}} comprises 206K fact-checking articles in 39 languages and 28K social media posts in 27 languages. Additionally, this dataset includes 31K pairs between social media posts and fact-checking articles, linking posts to corresponding fact-checking articles. Moreover, each fact-checking article is assigned a veracity rating and contains a URL, allowing retrieval of the full article content. This supplementary information enhances our pipeline by enabling a more structured and comprehensive evaluation of detecting previously fact-checked claims. 

\paragraph{AFP-Sum.}

To evaluate the abilities of LLMs to summarize fact-checking articles, we created the \textbf{\textit{AFP-Sum}} dataset, consisting of fact-checking articles and their summaries from the AFP (\textit{Agence France-Presse})\footnote{\url{https://www.afp.com}}. We scrapped fact-checking articles across 23 languages, yielding approximately 19K fact-checking articles with summaries written by fact-checkers. For our experiments, we selected 100 fact-checking articles per language, evaluating LLM-generated summaries in English. To facilitate evaluation, we employed Google Translate to translate all reference summaries into English. Table~\ref{tab:summary_dataset} in Appendix~\ref{app:afp-dataset} presents the dataset statistics.

\subsection{Language Models}
\label{sec:models}

We employed two categories of models, especially \textit{text embedding models} and \textit{large language models}. TEMs were used in the retrieval stage to identify the most relevant fact-checks for a given input. While numerous TEMs exist, we selected both English and multilingual models, using BM25 as a baseline for comparison. The TEMs used in our study are listed in Table~\ref{tab:retrieval-results}.

In addition to TEMs, we evaluated a diverse set of LLMs, including both closed and open-sourced, chosen based on their state-of-the-art performance across NLP tasks. For the summarization, we also experimented with smaller LLMs with fewer than 3 billion parameters. The full list of LLMs used in our experiments is shown Table~\ref{tab:model}.

\begin{table}
\resizebox{\columnwidth}{!}{%
\begin{tabular}{@{}llrll@{}}
\toprule
\textbf{Model} & \textbf{\# Params [B]} & \multicolumn{1}{l}{\textbf{\# Langs}} & \textbf{Organization} & \textbf{Citation} \\ \midrule
\texttt{GPT-4o} (2024-08-06) & N/A & \multicolumn{1}{r}{N/A} & OpenAI &  \\
\texttt{Claude 3.5 Sonnet} & N/A & \multicolumn{1}{r}{N/A} & Anthropic &  \\ \midrule
\texttt{Mistral Large} & 123 & 11 & Mistral AI & \citet{mistral-large} \\
\texttt{C4AI Command R+} & 104 & 23 & Cohere For AI & \citet{cohere_for_ai_2024} \\
\texttt{Qwen2 Instruct} & 72 & 29 & Alibaba & \citet{yang2024qwen2technicalreport} \\
\multirow[t]{4}{*}{\texttt{Qwen2.5 Instruct}} & 0.5, 1.5, 3, 72 & \multirow[t]{4}{*}{29} & \multirow[t]{4}{*}{Alibaba} & \multirow[t]{4}{*}{\citet{yang2024qwen2technicalreport}} \\
\texttt{Llama3.1 Instruct} & 70 & 8 & Meta & \citet{grattafiori2024llama3herdmodels} \\
\multirow[t]{2}{*}{\texttt{Llama3.2 Instruct}} & 1, 3 & \multirow[t]{2}{*}{8} & \multirow[t]{2}{*}{Meta} & \citet{grattafiori2024llama3herdmodels} \\
\multirow[t]{2}{*}{\texttt{Llama3.3 Instruct}} & 70 & \multirow[t]{2}{*}{8} & \multirow[t]{2}{*}{Meta} & \citet{grattafiori2024llama3herdmodels} \\
\multirow[t]{2}{*}{\texttt{Gemma3}} & 27 & \multirow[t]{2}{*}{140} & \multirow[t]{2}{*}{Google} & \citet{gemmateam2025gemma3technicalreport} \\
\bottomrule
\end{tabular}
}
\caption{LLMs used in our experiments, including both closed-source and open-source models.}
\label{tab:model}
\end{table}

\subsection{Evaluation}
\label{sec:eval}

We employed various evaluation metrics tailored to different stages of our proposed pipeline.

In the retrieval step, we used \textit{success-at-K} (S@K) as the primary evaluation metric for assessing TEMs performance. S@K measures the percentage of cases where a correct fact-check appears within the top \textit{K} retrieved results. Additionally, we apply this metric to evaluate the ability of LLMs to identify the most relevant fact-checks from the set retrieved by a TEM. 

For summarization experiments, we used two standard metrics: \textit{BERTScore} and \textit{ROUGE-L}. BERTScore evaluates semantic similarity by computing the F1 score based on contextual word embeddings from a BERT model. ROUGE, on the other hand, measures n-gram overlaps between the generated summary and the reference summary. Specifically, we employed ROUGE-L, which focuses on the longest common subsequence of words. ROUGE-L also helps detect cases where the LLM generated summaries in a language other than English -- something that is more challenging to identify using BERTScore.

Finally, for veracity prediction experiments, we employed standard classification metrics for imbalanced data: Macro F1 score, Precision and Recall.

\section{Retrieval Experiments}
\label{sec:retrieval}

In this section, we describe experiments with various TEMs in two settings. First, \textit{direct retrieval} (Section~\ref{sec:simple-retrieval}), in which we evaluate the performance of existing TEMs for retrieving the most similar previously fact-checked claims based on posts. Second, \textit{criteria-based retrieval} (Section~\ref{sec:interactive-retrieval}), where we evaluate TEMs for filtering the results based on criteria specified in a natural language in English (e.g. the presence of a specific named entity, the publication date, etc.).

\subsection{Direct Retrieval}
\label{sec:simple-retrieval}

We evaluated various TEMs and their performance for ranking previously fact-checked claims based on a given input. We formulate the task as a ranking problem, where we aim to rank all fact-checks from the database for a given input based on cosine similarity. We selected 20 languages with at least 100 posts per language with a setup similar to that proposed by \citet{pikuliak-etal-2023-multilingual}. In addition to the TEMs evaluated in~\cite{pikuliak-etal-2023-multilingual}, we included more recent multilingual TEMs, especially multilingual E5 models of various sizes. We evaluate the TEM's performance only in a monolingual setting, where fact-checked claims and posts are in the same language.

\paragraph{Results.}

The results of TEMs in the monolingual setting are shown in Table~\ref{tab:retrieval-results}. These results demonstrated that \textbf{some multilingual TEMs can outperform the combination of English translation with English TEMs}, but not statistically significantly. \texttt{Multilingual E5 Large} achieved the highest S@10 (statistically significant; $p<0.05$), while \texttt{GTR-T5-Large} achieved the best results with English translations ($p<0.05$). The other multilingual TEMs fall behind the English TEMs.

Table~\ref{tab:retrieval-languages} shows the results of all studied TEMs across 20 languages. Based on the results of the English TEMs, \texttt{GTR-T5-Large} achieved superior performance for most languages ($p<0.05$). However, for the German language, the results were lower than 0.70. On the other hand, \texttt{Multilingual E5 Large} proved to be effective across all languages ($p<0.05$), except Thai, where the smaller \texttt{Multilingual E5} outperformed larger versions.

\begin{table}[t]
\resizebox{\columnwidth}{!}{%
\begin{tabular}{lrcr}
\toprule
\textbf{Model} & \textbf{Size [M]} & \textbf{Ver.} &\textbf{Avg. S@10} \\
\midrule
BM25 &  & Og & 0.62 \\ \midrule
\multicolumn{3}{c}{\textit{\textbf{English TEMs}}} \\\midrule
\texttt{DistilRoBERTa} & 82 & En & 0.75 \\
\texttt{MiniLM-L6} & 22 & En & 0.79 \\
\texttt{MiniLM-L12} & 33 & En & 0.79 \\
\texttt{MPNet-Base} & 109 & En & 0.77 \\
\texttt{GTE-Large-En} & 434 & En & 0.80 \\
\texttt{GTR-T5-Large} & 737 & En & \textbf{0.83} \\\midrule
\multicolumn{3}{c}{\textit{\textbf{Multilingual TEMs}}} \\\midrule
\texttt{BGE-M3} & 568 & Og & 0.82 \\
\texttt{DistilUSE-Base-Multilingual} & 134 & Og & 0.66 \\
\texttt{LaBSE} & 470 & Og & 0.69 \\
\texttt{Multilingual E5 Small} & 117 & Og & 0.78 \\
\texttt{Multilingual E5 Base} & 278 & Og & 0.78 \\
\texttt{Multilingual E5 Large} & 559 & Og & \textbf{0.84} \\
\texttt{MiniLM-L12-Multilingual} & 117 & Og & 0.63 \\
\texttt{MPNet-Base-Multilingual} & 278 & Og & 0.69 \\
\bottomrule
\end{tabular}
}
\caption{Average performance of English and multilingual TEMs in monolingual settings using S@10 metric. Ver. denotes either \textit{original} (Og) or the \textit{English} (En) version of the dataset. The best results for both versions are highlighted in \textbf{bold}.}
\label{tab:retrieval-results}
\end{table}

\subsection{Criteria-based Retrieval}
\label{sec:interactive-retrieval}

In addition to retrieval based only on the input claim or posts, we also experimented with criteria-based retrieval, where we employ TEMs to filter the results based on given criteria, e.g., the presence of a specific named entity. The aim is to evaluate whether TEMs can be employed to filter the results with natural language instructions provided by fact-checkers. We defined four settings for the experiments: filtering based on the \textit{language}, \textit{date range}, \textit{fact-checking domain} or the \textit{named entity}. We selected the best-performing TEM -- \texttt{Multilingual E5 Large} for the experiments. We proposed a template illustrated in Figure~\ref{fig:template}, consisting of fact-checked claims and metadata, such as language, fact-checking organization, and publish date.

As ground truth, we used the results obtained by using \texttt{Multilingual E5 Large} to rank a subset of the data already filtered based on a given condition using the manually-designed filter (e.g., only Spanish fact-checks were ranked). Our pipeline for criteria-based retrieval consists of two steps: \textit{(1) Retrieval based on the criteria} (e.g., a given language), where we select only fact-checks with a similarity score of more than 0.8; \textit{(2) Ranking based on the post}, where we rank previously retrieved results using the post content, similarly to the direct retrieval.

For the evaluation, we employed \textit{Spearman's rank correlation coefficient} and \textit{Kendall's Tau} to evaluate the capabilities of TEMs to rank the results by using queries in the natural language. We calculated the correlation between ranks produced by \texttt{Multilingual E5 Large} with our two-step approach and ranks obtained by using the \texttt{Multilingual E5 Large} on the already filtered results based on manually-designed filters. 

\paragraph{Results.}

Table~\ref{tab:interactive-retrieval} presents the results for the filtered retrieval across four settings: named entities, languages, fact-checking domains and date ranges. We calculated the average Spearman's rank correlation, Kendall's Tau and the proportion of the common fact-checks between the predicted and reference list of fact-checks. A positive correlation indicates that the predicted ranking aligns with the reference ranking, whereas a negative correlation suggests an inverse relationship. 

Our results showed that \textbf{filtering based on the named entities yielded the highest overlap between the predicted and ground truth fact-check lists} ($p<0.05$), suggesting that TEMs performed best when fact-checks were retrieved based on named entities. Despite this, the Spearman correlation of $-0.31$ indicates that while TEMS might identify relevant fact-checks, their rankings did not fully match the ground truth ordering.

\begin{table}[t]
\resizebox{\columnwidth}{!}{%
\begin{tabular}{lrrr}
\toprule
\multicolumn{1}{c}{\textbf{Settings}} & \multicolumn{1}{c}{\textbf{\begin{tabular}[c]{@{}c@{}}Avg.\\ Spearman\end{tabular}}} & \multicolumn{1}{c}{\textbf{\begin{tabular}[c]{@{}c@{}}Avg.\\ Kendall's Tau\end{tabular}}} & \multicolumn{1}{c}{\textbf{\begin{tabular}[c]{@{}c@{}}Avg.\\ Common FCs\end{tabular}}} \\
\midrule
Named Entities & $-0.31$ & $-0.20$ & $0.32$ \\
Languages & $-0.58$ & $-0.43$ & $0.17$ \\
Domains & $-0.66$ & $-0.51$ & $0.12$ \\
Dates & $-0.82$ & $-0.64$ & $0.02$ \\ 
\bottomrule
\end{tabular}
}
\caption{Scores for average \textit{Spearman correlation coefficient}, \textit{Kendall's Tau} and the \textit{proportion of the common fact-checks} (FCs) between the ground truth and predicted ranked list. We report the mean score across all settings with at least 100 fact-checks per category.}
\label{tab:interactive-retrieval}
\end{table}

Filtering by language, domain and date range led to lower performance, with the latter performing the worst. This suggests that while TEMs can retrieve relevant fact-checks based on natural language instructions, their filtering changes the candidate set that limits and reduces overall ranking performance. Additionally, we specified a date range, whereas the embeddings of fact-checks only included the exact date of each fact-check. This discrepancy made it more challenging for TEMs to retrieve the fact-checks based on dates not explicitly mentioned in the prompt.

\section{Filtration Experiments}
\label{sec:filtration}

To filter out irrelevant previously fact-checked claims, we experimented with several LLMs on a subset of the \textit{\textbf{MultiClaim}} dataset. We selected 10 languages, specifically \textit{Czech}, \textit{English}, \textit{French}, \textit{German}, \textit{Hindi}, \textit{Hungarian}, \textit{Polish}, \textit{Portuguese}, \textit{Spanish} and \textit{Slovak}, with 100 posts per language. These posts were chosen based on their veracity labels. However, since the \textbf{\textit{MultiClaim}} dataset predominantly contains false posts, achieving a balanced distribution was not feasible. The final dataset consists of 55 true, 65 unverifiable and 880 false posts, resulting in a significant imbalance. We used this data to evaluate the efficiency of LLMs in filtering out irrelevant fact-checks for a given input. 

Our approach involved a two-step process. First, we used \texttt{Multilingual E5 Large} to retrieve the 50 most similar fact-checked claims. Then, we instructed the LLM to filter this set (see Figure~\ref{fig:pipeline_prompts}), selecting only those directly relevant to the input post, while removing irrelevant fact-checks.

To assess the performance and efficiency of the LLMs in this task, we calculated the S@10 and MRR (mean reciprocal rank) scores for retrieval. In addition, we calculated Macro F1, True negative rate (TNR) and False negative rate (FNR) to identify the capabilities of LLMs. To calculate classification metrics, we created pairs of posts and fact-checks identified by the \texttt{Multilingual E5 Large} model, where the relevance labels were obtained from the labelled pairs from the \textit{\textbf{MultiClaim}} dataset. TNR represents the proportion of how many irrelevant fact-checks were correctly filtered out, while FNR represents the proportion of how many relevant fact-checks were incorrectly filtered out. In this case, we want to maximize the TNR and minimize the FNR.  

\subsection{Results}

\begin{table}
\resizebox{\columnwidth}{!}{%
\begin{tabular}{lrrrrr}
\toprule
\textbf{Model} & \textbf{S@10} $\uparrow$ & \textbf{MRR} $\uparrow$ & \textbf{Macro F1} $\uparrow$ & \textbf{TNR} $\uparrow$ & \textbf{FNR} $\downarrow$\\
\midrule
\texttt{Multilingual E5 Large} & \textbf{0.76} & \textbf{0.58} & 54.75 & 86.27 & 25.59\\
\midrule
\texttt{Mistral Large 123B} & \underline{0.70} & \underline{0.40} & \underline{59.82} & 90.23 & \textbf{15.38} \\
\texttt{C4AI Command R+} & 0.66 & 0.35 & 55.50 & 85.83 & \textbf{15.38} \\
\texttt{Qwen2.5 72B} & 0.57 & 0.32 & 58.37 & 90.81 & 30.65 \\
\texttt{Llama3.3 70B} & 0.67 & 0.38 & \textbf{59.96} & 90.82 & \underline{19.61} \\
\texttt{Llama3.1 70B} & 0.63 & 0.37 & 59.62 & \underline{91.08} & 24.25 \\
\texttt{Gemma3 27B} & 0.65 & 0.35 & 57.77 & 89.14 & 21.78 \\
\texttt{Llama3.1 8B} & 0.60 & 0.24 & 52.38 & 82.30 & 21.16 \\
\texttt{Qwen2.5 7B} & 0.47 & 0.35 & 59.25 & \textbf{93.20} & 43.86 \\ \bottomrule
\end{tabular}
}
\caption{Retrieval and filtration performance results on 100 posts across 10 languages. \texttt{Multilingual E5 Large} serves as the baseline. The best results are highlighted in \textbf{bold}, while the second-best results are \underline{underlined}.}
\label{tab:filtration}
\end{table}

Table~\ref{tab:filtration} summarizes our results on filtering irrelevant fact-checks. Using \texttt{Multilingual E5 Large} as the baseline, we correctly retrieved 76\% of relevant fact-checks within the top 10 results. To further assess performance, we framed the ranking task as binary classification, selecting an optimal threshold using Youden's Index. Macro F1 showed that the baseline outperformed \texttt{Llama3.1 8B}.

After retrieval, we applied an LLM to filter the top 50 retrieved fact-checks. While this lowered S@10 and MRR scores compared to the baseline, the aim was to reduce the number of irrelevant fact-checks presented to fact-checkers. We measured the proportion of relevant and irrelevant fact-checks removed. \textbf{\texttt{Mistral Large} achieved the best trade-off between TNR and FNR} ($p<0.05$), while also outperforming other LLMs in S@10.

Our findings suggest that \textbf{while LLMs effectively remove irrelevant fact-checks, they may exclude some relevant ones}. The performance gap between \texttt{Multilingual E5 Large} and LLMs indicates occasional misclassification of relevant fact-checks as irrelevant, although LLMs may also elevate lower-ranked fact-checks into the top 10.

\section{Summarization Evaluation}
\label{sec:summarization}

We evaluated LLMs on summarizing fact-checking articles using a subset of our \textbf{\textit{AFP-Sum}} dataset across 23 languages. Experiments were conducted in two settings: (1) \textit{Article first} -- the article is provided before the instruction; (2) \textit{Article last} -- the article is provided after the instruction (see Figure~\ref{fig:summary_prompts}). We examined how prompt order and quantization affect summary quality. Articles were provided in their original language, with instructions to generate a summary in English. The generated summaries were compared against English translations of reference summaries using Google Translate.

\begin{figure}
    \centering
    \includegraphics[width=\columnwidth]{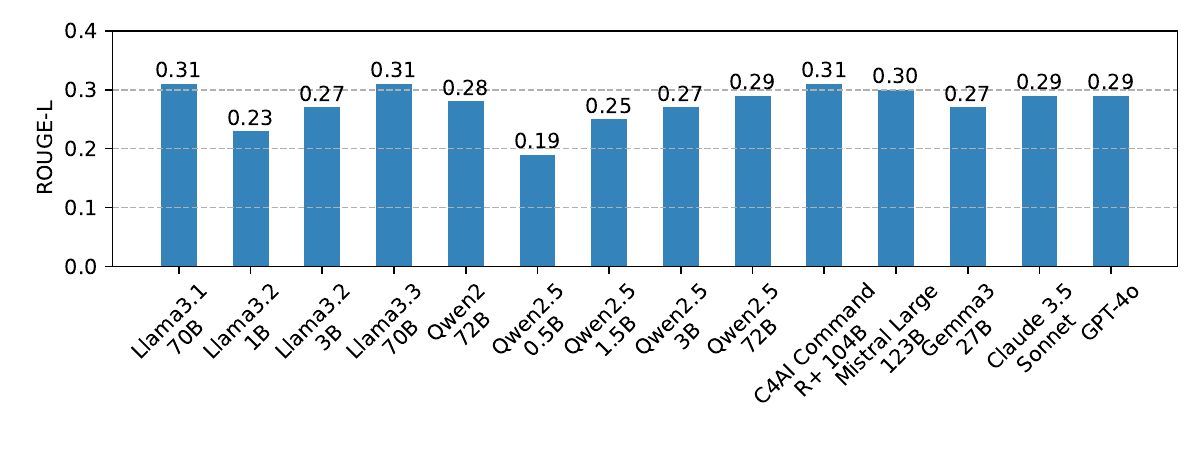}
    \caption{Overall performance of LLMs for fact-check summarization. We report the average ROUGE-L score for each LLM using the \textit{Article first} setup, where the article is provided before the instruction.}
    \label{fig:summary-results}
\end{figure}

\subsection{Results}

Figure~\ref{fig:summary-results} presents the overall results using the ROUGE-L metric in the \textit{Article first} setup. The results demonstrated the diverse performance across LLMs. \textbf{Smaller \texttt{Llama} models struggled with summarization, often generating summaries in the article's original language} instead of English, leading to lower ROUGE-L. In contrast, other LLMs better adhered to the instruction to produce English summaries. Furthermore, providing the instruction before the article worsened this issue and resulted in a very low ROUGE-L (see Table~\ref{tab:rouge-quant}).

Table~\ref{tab:summary-quantization} compares \texttt{Llama3.2} models (1B and 3B parameters) across different article-order setups and three quantization levels. Summaries were generated using 4-bit, 8-bit, and full-precision models. Results showed that \textbf{providing the article before the instruction significantly improved performance} ($p<0.05$), yielding better results when the article was provided after the instruction. For \texttt{Llama3.2 1B}, the 8-bit model generally performed best (not statistically significant), with full precision close behind. The performance gap between the full-precision and 4-bit was around 0.3 BERTScore points for \texttt{Llama3.2 1B}, while for \texttt{Llama3.2 3B}, only 0.1, suggesting that \textbf{quantization has less impact on larger LLMs}.

Overall, \texttt{Llama3.3 70B} and \texttt{Mistral Large} achieved the best performance across languages (see Table~\ref{tab:summary-rouge}), while other LLMs lagged behind (not statistically significant). The results indicate that LLMs covering fewer languages (e.g., \texttt{Llama}) can outperform broader multilingual LLMs, e.g., \texttt{C4AI Command R+} or \texttt{Qwen2.5}.

\section{Evaluation of LLM's Veracity Prediction}
\label{sec:veracity}

To assess how well LLMs predict claim veracity using retrieved previously fact-checked claims and the fact-check summaries (see Figure~\ref{fig:pipeline_prompts}), we employed the same data as in Section~\ref{sec:filtration}. The final dataset consists of three classes: \textit{True}, \textit{False} or \textit{Unverifiable}, which are imbalanced. Therefore, we leveraged Macro F1, Macro Precision and Macro Recall to evaluate the performance of LLMs. In this case, the supplementary information is in English, particularly summaries, ratings and fact-checked claims. This is also beneficial for human fact-checkers to understand the results provided by LLMs. As baselines, we selected \texttt{Mistral Large} and \texttt{Llama3.3}, instructed only with the post and task description without additional information.

\subsection{Results}

\begin{table}
\centering
\small
\begin{tabular*}{\columnwidth}{@{\extracolsep{\fill}}lccc@{}}
\toprule
\multicolumn{1}{c}{\textbf{Model}} & \multicolumn{1}{c}{\textbf{\begin{tabular}[c]{@{}c@{}}Macro\\F1\end{tabular}}} & \multicolumn{1}{c}{\textbf{\begin{tabular}[c]{@{}c@{}}Macro\\Precision\end{tabular}}} & \multicolumn{1}{c}{\textbf{\begin{tabular}[c]{@{}c@{}}Macro\\Recall\end{tabular}}} \\
\midrule
\multicolumn{4}{c}{\textit{Baseline (without retrieved fact-checks)}} \\
\midrule
\texttt{Mistral Large 123B} & 26.53 & 39.57 & 33.25 \\
\texttt{Llama3.3 70B} & 30.29 & 34.22 & 33.30 \\
\midrule
\texttt{Mistral Large 123B} & \textbf{63.05} & \textbf{64.88} & \textbf{61.62} \\
\texttt{C4AI Command R+} & 54.92 & 55.50 & 54.38 \\
\texttt{Qwen2.5 72B} & 57.28 & 57.28 & 57.33 \\
\texttt{Llama3.3 70B} & 52.62 & 52.18 & 53.09 \\
\texttt{Llama3.1 70B} & 51.68 & 50.67 & 53.25 \\
\texttt{Gemma3 27B} & 52.39 & 52.62 & 52.36 \\
\texttt{Llama3.1 8B} & 49.15 & 46.66 & 53.65 \\
\texttt{Qwen2.5 7B} & 51.99 & 56.47 & 49.23 \\ \bottomrule
\end{tabular*}
\caption{Veracity prediction results across various LLMs. Results are presented for baseline (without retrieved fact-checks), and LLMs with retrieved fact-checks. The best results are highlighted in \textbf{bold}.}
\label{tab:veracity}
\end{table}

Our results are shown in Table~\ref{tab:veracity}, where we employed eight LLMs with different model sizes. The \textbf{\texttt{Mistral Large} with the retrieved information outperformed all other LLMs}, also the baselines (not statistically significant). It achieved the highest performance, making it the most reliable for veracity prediction out of the experimented LLMs.

\texttt{Qwen2.5 72B} follows with a noticeable drop in performance, suggesting that model size alone does not determine effectiveness. \texttt{Llama} models performed similarly, showing limited ability to distinguish veracity class based on the retrieved information. \textbf{The smaller models performed the worst and struggled with generalization.}

Overall, while bigger LLMs tend to perform better, \textbf{the contextual information plays an important role}. The strong performance of \texttt{Mistral Large} highlights its potential for improving fact-checking applications.

\section{Human Evaluation}

To assess the effectiveness of our proposed pipeline for multilingual claim retrieval, we developed a web-based tool designed for fact-checkers. In addition to conducting automatic evaluations of individual components, we focused on human evaluation of the entire pipeline using the developed tool. We provided the tool to students and academics, who assessed its performance and usability. Their feedback was collected through the evaluation workshop and a structured questionnaire, offering insights into the system's applicability.

\subsection{Developed Tool}

Our web-based application\footnote{\url{https://fact-exu-miner.kinit.sk/}} integrates the pipeline described in Section~\ref{sec:methodology}, utilizing the best-performing TEM model -- \texttt{Multilingual E5 Large}. The backend employs \texttt{Llama3.3 70B}, selected for its strong summarization ability and ability to filter irrelevant fact-checks. We store fact-checked claims from over 80 languages, along with metadata and \texttt{Multilingual E5 Large} embeddings, in the Milvus\footnote{\url{https://github.com/milvus-io/milvus}} vector database. The result of the system provides a ranked list of relevant fact-checks identified by the LLM, along with their summaries and explanations. In addition, we provide users with a veracity label distribution graph and a verdict explanation to aid decision-making.

\subsection{Evaluation \& Results}

To evaluate the tool, we conducted a user study with six participants (five journalism students and one academic). Each participant interacted with the tool and completed the questionnaire designed to assess system usability, output quality and overall effectiveness in supporting fact-checking.

Participants rated their satisfaction with various aspects of the tool, including summaries, explanations of relevant fact-checks, the veracity graph and overall usability. Ratings ranged from 1 (very unsatisfied) to 5 (very satisfied). Most features received average satisfaction scores between 3.5 and 4, with explanations of relevant fact-checks and veracity explanations achieving the highest average rating. Users generally found the tool helpful in identifying relevant fact-checks and appreciated the clarity of summaries and explanations.

\begin{figure}
    \centering
    \includegraphics[width=\columnwidth]{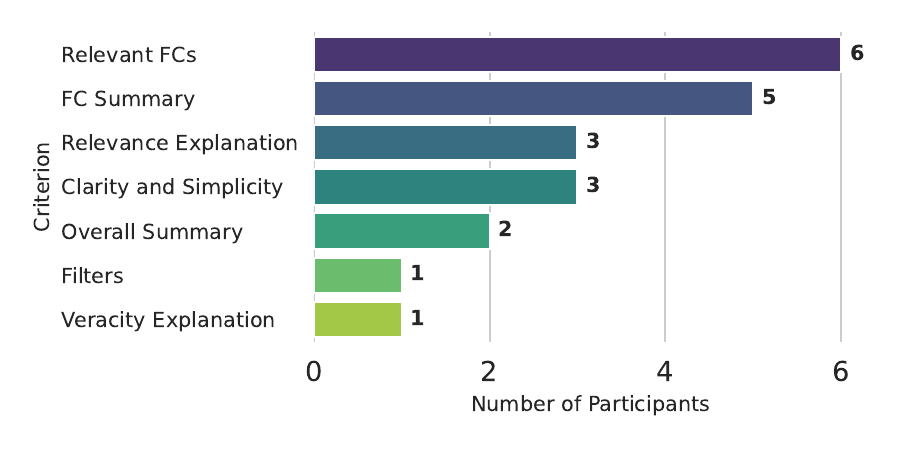}
    \caption{Number of participants ($N=6$) who highlighted each evaluation criterion as beneficial.}
    \label{fig:human-eval}
\end{figure}

We asked participants to assess the tool's main benefits (see Figure~\ref{fig:human-eval}). Participants highlighted the retrieval of relevant fact-checks (FCs), concise summaries and explanations, and the clarity of the interface as the main benefits. These results suggest that the tool is a promising aid for fact-checkers.

\section{Discussion}

\paragraph{Multilingual TEMs Outperform English TEMs.}

\texttt{Multilingual E5 Large} achieved the best retrieval performance across most languages. However, \textbf{criteria-based retrieval} experiments showed that TEMs struggled with natural language instructions, especially when filtering by date range.

\paragraph{Filtration with LLMs Improves Precision But Has Trade-Offs.}

\texttt{Mistral Large} provided the best balance between retaining relevant fact-checks and filtering out irrelevant ones, showing promise for assisting fact-checkers. However, the trade-off between precision and recall remains challenging, as some useful fact-checks may be excluded.

\paragraph{Larger LLMs Excel in Summarization and Veracity Prediction.}

Smaller LLMs often failed to follow instructions, producing summaries in the original language instead of English. Larger LLMs performed better, particularly when the article preceded the instruction, and were more effective at predicting claim veracity. However, overall performance remained moderate due to the inherent difficulty of accurately assessing claim veracity.

\section{Conclusion}

This paper presents a pipeline for multilingual retrieval of previously fact-checked claims, integrating LLMs to enhance the fact-checking process. Beyond the retrieval, our approach supports fact-checkers by filtering irrelevant fact-checks, summarizing fact-checking articles, and predicting veracity labels along with explanations. We also developed a web-based application and evaluated its effectiveness in the fact-checking process. Our findings demonstrate the potential of LLMs to improve fact-checking workflows, making them more efficient and accessible across languages.

\section*{Limitation}

\paragraph{Models Used.}

Our experiments relied on a selection of state-of-the-art LLMs and TEMs, including closed-source models (e.g., \texttt{GPT-4o} or \texttt{Claude 3.5 Sonnet}) and open-sourced models (e.g., \texttt{Mistral Large}, \texttt{Llama3}). However, model performance is highly dependent on training data and fine-tuning strategies. As a result, our findings may not generalize to all LLMs and architectures, and future improvements may arise with newer models.

\paragraph{Language Support.}

Despite evaluating our approach on more than 10 languages and incorporating fact-checking data from 20 languages, our system may still face challenges in handling low-resource languages. The performance of TEMs and LLMs may vary across languages, particularly those with limited pre-trained resources. 

In addition, the selected LLMs exhibit varying degrees of multilingual capabilities. While model cards on Hugging Face\footnote{\url{https://huggingface.co/}} indicate intended language support, the models may demonstrate capabilities in additional languages due to the training data diversity and potential data contamination.

\paragraph{Human Evaluation.}

Our user study included six participants from an academic environment -- five journalism students and one academic. While professional fact-checkers would have been more appropriate evaluators for our tool, their inclusion was not feasible due to time constraints and limited availability. Journalism students, however, serve as a reasonable proxy, given their specialization and relevance as potential end-users. We acknowledge this limitation and consider evaluation with professional fact-checkers as an important direction for future work.

\paragraph{Automated Veracity Prediction.}

Our pipeline includes an LLM-based veracity prediction, which suggests a claim's veracity based on retrieved fact-checks. However, automated assessments remain limited by the availability and accuracy of fact-checking data. In cases where no relevant fact-check exists, the system may struggle to provide reliable predictions.

\section*{Ethical Consideration}

\paragraph{Biases.}

Since we experimented with LLMs, our system may inherit biases from the training data used in the embedding models and LLMs. These biases can affect claim retrieval, relevant fact-check selection, and veracity assessments, potentially leading to skewed or misleading outputs, especially for politically sensitive or controversial topics.

Additional bias is propagated by fact-checkers since they decide what they will fact-check.

\paragraph{Developed Tool.}

The final version of the developed tool employs the \texttt{Llama3.3 70B}\footnote{\url{https://huggingface.co/meta-llama/Llama-3.3-70B-Instruct}}. This model was selected for its advanced capabilities in summarizing and efficient inference, compared to larger models. The tool includes biases inherited from the used LLM.

To enhance transparency and assist users in evaluating the output, the tool also provides the number of supporting and refuting fact-checks associated with a given claim. This information is included in the veracity distribution graph within the tool. The user can employ this information for the final decision on the veracity of the given claim and compare the veracity prediction done by the LLM.

The classification accuracy and efficiency of the pipeline depend on the final model -- in our case, \texttt{Llama3.3 70B}. The evaluation of the corresponding model and its effectiveness for the veracity prediction is evaluated in Section~\ref{sec:veracity}. LLMs are known to hallucinate~\cite{rawte2023surveyhallucinationlargefoundation}, and therefore, they might create fake, non-factual or even harmful information.

In addition, the tool incorporates fact-checking articles and corresponding claims, many of which are false or misleading statements spread online. As a result, users of the application may be exposed to false, misleading, or even harmful claims. To address this, the tool includes a Terms of Use that outlines its intended purpose, identifies the target users, and specifies user groups for whom the tool is not intended.

\paragraph{Personal Information.}

The original \textit{\textbf{MultiClaim}}~\cite{pikuliak-etal-2023-multilingual} dataset might contain personal information and data from the social media posts (e.g., the names of users). However, we are not using any personal information within our experiments or the developed tool.

\paragraph{Terms of Use of Platforms and Datasets.}

In our research, we utilized the \textit{\textbf{MultiClaim}} dataset~\cite{pikuliak-etal-2023-multilingual}, which is accessible under specific conditions -- the dataset is restricted to academic and research purposes.

Additionally, we incorporated fact-checking articles from Agence France-Presse (AFP)\footnote{\url{https://factcheck.afp.com/}}, which are available for personal, private, and non-commercial use. Any reproduction or redistribution beyond these permitted uses is forbidden. 

We ensure that our use of both the \textbf{\textit{MultiClaim}} and AFP's content for the \textit{\textbf{AFP-Sum}} is compiled with their respective terms and conditions.

\paragraph{Intended Use.}

The annotated data presented in this research are intended solely for research purposes. They are derived from the existing \textbf{\textit{MultiClaim}} dataset~\cite{pikuliak-etal-2023-multilingual}, which is also intended only for research purposes. In our work, we selected a subset and annotated specific portions for the task of veracity prediction. 

Additionally, we introduce the \textbf{\textit{AFP-Sum}} dataset, comprising fact-checking articles and their summaries sourced from the AFP organization. Due to the copyright restrictions on the AFP data, its usage is strictly limited to research purposes. As such, we release the \textbf{\textit{AFP-Sum}} dataset and any derived resources to researchers for non-commercial research use only.

To promote reproducibility, we also release code used to obtain our results. Both the datasets and the code are intended only for research use, and replicating our findings requires access to the original \textbf{\textit{MultiClaim}} dataset, which is available under its respective terms and conditions.

\paragraph{Usage of AI assistants.}

We have used the AI assistant for grammar checks and sentence structure improvements. We have not used AI assistants in the research process beyond the experiments detailed in the Methodology section (Sec. \ref{sec:methodology}). 

\section*{Acknowledgments}

This project is funded by the European Media and Information Fund (grant number 291191). The sole responsibility for any content supported by the European Media and Information Fund lies with the author(s) and it may not necessarily reflect the positions of the EMIF and the Fund Partners, the Calouste Gulbenkian Foundation and the European University Institute.

This work was supported by the Ministry of Education, Youth and Sports of the Czech Republic through the e-INFRA CZ (ID:90254).

\bibliography{custom}

\appendix

\section{Computational Resources}

For our experiments, we leveraged a computational infrastructure consisting of A40 PCIe 40GB, A100 80GB and H100 NVL 94GB NVIDIA GPUs. In addition, we used API from Anthropic to run the experiments with Claude 3.5 Sonnet and Azure for deploying GPT-4o.

The experiments with TEMs took around 30 GPU hours. Our experiments with summarization and comparison of various quantization variants required approximately 600 GPU hours. Finally, the experiments with the overall pipeline -- with veracity prediction -- took around 400 GPU hours.

\section{Dataset Statistics}

\subsection{MultiClaim Dataset}
\label{app:multiclaim}

For our experiments, we selected \textbf{\textit{MultiClaim}}~\cite{pikuliak-etal-2023-multilingual}, the most comprehensive multilingual dataset for previously fact-checked claim retrieval. We used the full dataset for retrieval experiments, as described in Section~\ref{sec:retrieval}. For other components, we worked with a subset of \textbf{\textit{MultiClaim}}, selecting a set of 10 languages that included both high- and low-resource languages. From each language, we sampled 100 social media posts for each language while aiming to balance the distribution of veracity labels. However, the original \textbf{\textit{MultiClaim}} dataset is highly imbalanced, with a predominant number of false social media posts. As a result, our subset contains a significant proportion of false claims. Table~\ref{tab:multiclaim-subset} provides detailed statistics on the subset used in our experiments.

The final veracity ratings were derived from fact-checking articles linked to particular posts. We manually evaluated these links to ensure they were correctly extracted from the metadata of the corresponding fact-checks.

\begin{table}
\centering
\resizebox{\columnwidth}{!}{%
\begin{tabular}{llrrrr}
\toprule
\textbf{Language} & \textbf{Lang. Code} & \textbf{\begin{tabular}[c]{@{}c@{}}Average WC\end{tabular}} & \textbf{\# False} & \textbf{\# True} & \textbf{\# Unverifiable} \\
\midrule
Czech & cs & 168.60 $\pm$ 242.44 & 100 & 0 & 0\\
German & de & 86.08 $\pm$ 84.90 & 94 & 1 & 5 \\
English & en & 111.11 $\pm$ 142.39 & 92 & 5 & 3 \\
French & fr & 109.14 $\pm$ 129.62 & 95 & 4 & 1 \\
Hindi & hi & 63.36 $\pm$ 108.82 & 95 & 4 & 1 \\
Hungarian & hu & 123.73 $\pm$ 178.21 & 97 & 0 & 3 \\
Polish & pl & 102.00 $\pm$ 130.70 & 96 & 2 & 2 \\
Portuguese & pt & 92.25 $\pm$ 176.08 & 76 & 6 & 18 \\
Slovak & sk & 126.59 $\pm$ 214.57 & 100 & 0 & 0 \\
Spanish & es & 95.73 $\pm$ 130.48 & 35 & 33 & 32 \\
\midrule
\textbf{Total} & & & 880 & 55 & 65 \\ 
\bottomrule
\end{tabular}
}
\caption{Statistics of a subset of \textbf{\textit{MultiClaim}} dataset used for experiments with filtration and veracity prediction. We provide the average word count (WC) with standard deviation and the number of false, true and unverifiable claims per language.}
\label{tab:multiclaim-subset}
\end{table}

\subsection{AFP-Sum Dataset}
\label{app:afp-dataset}

To assess the ability of LLMs to summarize fact-checking articles, we collected data from the AFP organization. Specifically, we extracted fact-checking articles in 23 languages, listed in Table~\ref{tab:afp-sum-stats}, which also includes the number of articles per language. Our dataset comprises fact-checking articles published up until September 2023.

\begin{table}[t]
\resizebox{\columnwidth}{!}{%
\begin{tabular}{lllr}
\toprule
\textbf{Language} & \textbf{Lang. Code} & \textbf{Domain} & \textbf{\# Articles} \\
\midrule
English & en & https://factcheck.afp.com & 6358 \\
Spanish & es & https://factual.afp.com & 3999 \\
French & fr & https://factuel.afp.com & 2883 \\
Portuguese & pt & https://checamos.afp.com & 1320 \\
German & de & https://faktencheck.afp.com & 564 \\
Indonesian & id & https://periksafakta.afp.com & 506 \\
Polish & pl & https://sprawdzam.afp.com & 386 \\
Korean & ko & https://factcheckkorea.afp.com & 359 \\
Thai & th & https://factcheckthailand.afp.com & 349 \\
Serbian & sr & https://cinjenice.afp.com & 306 \\
Finnish & fi & https://faktantarkistus.afp.com & 289 \\
Malay & ms & https://semakanfakta.afp.com & 233 \\
Slovak & sk & https://fakty.afp.com & 226 \\
Czech & cs & https://napravoumiru.afp.com & 216 \\
Dutch & nl & https://factchecknederland.afp.com & 192 \\
Bulgarian & bg & https://proveri.afp.com & 139 \\
Bengali & bn & https://factcheckbangla.afp.com & 136 \\
Romanian & ro & https://verificat.afp.com & 135 \\
Burmese & my & https://factcheckmyanmar.afp.com & 128 \\
Hindi & hi & https://factcheckhindi.afp.com & 125 \\
Greek & el & https://factcheckgreek.afp.com & 121 \\
Hungarian & hu & https://tenykerdes.afp.com & 112 \\
Catalan & ca & https://comprovem.afp.com & 110 \\
\bottomrule
\end{tabular}
}
\caption{Statistics of the \textbf{\textit{AFP-Sum}} dataset, consisting of the languages, language codes, domains and the number of articles per language.}
\label{tab:afp-sum-stats}
\end{table}

For the final evaluation, we employed only a subset of the data, especially we used 100 fact-checking articles per language, which we randomly sampled from the \textbf{\textit{AFP-Sum}} dataset. The statistics of the sampled dataset, consisting of 2300 fact-checking articles in 23 languages, are shown in Table~\ref{tab:summary_dataset}. Besides the number of fact-checking articles, we provide the average word count for the article and for the summary along with the standard deviation.

Since the extracted summaries are in the original language, we employed Google Translate API to translate the summaries into English, which we then used for the final evaluation and calculating the BERTScore and ROUGE-L.

\begin{table}[t]
\resizebox{\columnwidth}{!}{%
\begin{tabular}{llrll}
\toprule
\multicolumn{1}{c}{\textbf{\begin{tabular}[c]{@{}c@{}}Lang. Code\end{tabular}}} & \multicolumn{1}{c}{\textbf{Language}} & \multicolumn{1}{c}{\textbf{\# Articles}} & \multicolumn{1}{c}{\textbf{\begin{tabular}[c]{@{}c@{}}Average WC\\Article\end{tabular}}} & \multicolumn{1}{c}{\textbf{\begin{tabular}[c]{@{}c@{}}Average WC\\Summary\end{tabular}}} \\
\midrule
bg & Bulgarian & 100 & 965.66 $\pm$ 533.28 & 81.57 $\pm$ 20.09 \\
bn* & Bengali & 100 & 308.93 $\pm$ 114.53 & 55.07 $\pm$ 17.23 \\
ca* & Catalan & 100 & 822.30 $\pm$ 454.67 & 82.69 $\pm$ 18.19 \\
cs & Czech & 100 & 691.35 $\pm$ 353.20 & 62.31 $\pm$ 14.28 \\
de & German & 100 & 869.32 $\pm$ 510.19 & 62.19 $\pm$ 15.57 \\
el & Greek & 100 & 1116.24 $\pm$ 500.74 & 86.51 $\pm$ 17.89 \\
en & English & 100 & 463.63 $\pm$ 197.19 & 58.18 $\pm$ 13.51 \\
es & Spanish & 100 & 713.13 $\pm$ 477.01 & 75.87 $\pm$ 18.69 \\
fi* & Finnish & 100 & 754.15 $\pm$ 369.82 & 57.50 $\pm$ 17.54 \\
fr & French & 100 & 659.96 $\pm$ 568.90 & 61.38 $\pm$ 23.46 \\
hi & Hindi & 100 & 507.20 $\pm$ 142.50 & 78.07 $\pm$ 17.16 \\
hu & Hungarian & 100 & 884.79 $\pm$ 570.36 & 78.02 $\pm$ 17.50 \\
id* & Indonesian & 100 & 458.79 $\pm$ 173.84 & 56.58 $\pm$ 12.63 \\
ko & Korean & 100 & 309.15 $\pm$ 131.40 & 46.99 $\pm$ 11.12 \\
ms & Malay & 100 & 521.20 $\pm$ 163.88 & 59.05 $\pm$ 13.16 \\
my & Burmese & 100 & 233.89 $\pm$ 77.18 & 31.18 $\pm$ 10.57 \\
nl & Dutch & 100 & 998.47 $\pm$ 515.52 & 73.51 $\pm$ 19.30 \\
pl & Polish & 100 & 836.52 $\pm$ 474.79 & 59.31 $\pm$ 17.34 \\
pt & Portuguese & 100 & 715.00 $\pm$ 343.31 & 80.21 $\pm$ 15.72 \\
ro & Romanian & 100 & 1156.78 $\pm$ 566.54 & 88.75 $\pm$ 19.20 \\
sk & Slovak & 100 & 850.55 $\pm$ 552.95 & 62.53 $\pm$ 22.07 \\
sr & Serbian & 100 & 954.83 $\pm$ 497.00 & 71.55 $\pm$ 19.63 \\
th & Thai & 100 & 121.34 $\pm$ 42.42 & 10.71 $\pm$ 4.68 \\
\bottomrule
\end{tabular}
}
\caption{Statistics of the dataset used for summarization experiments, consisting of 100 fact-check articles across 23 languages. Languages marked with * are not included in other experiments besides summarization. The Arabic language is missing, which is used in other experiments.}
\label{tab:summary_dataset}
\end{table}

\section{Retrieval Experiments}

Table~\ref{tab:retrieval-languages} provides the results of the experiments with simple retrieval across 20 languages, where we aimed to evaluate how accurate TEMs are for retrieving the relevant fact-checks based on the content of the social media post. We report S@10 as the main metric for the evaluation.

\begin{table*}
\resizebox{\textwidth}{!}{%
\begin{tabular}{lllllllllllllllllllllll}
\toprule
\multicolumn{1}{c}{\textbf{Model}} & \multicolumn{1}{c}{\textbf{Ver.}} & \multicolumn{1}{c}{\textbf{ara}} & \multicolumn{1}{c}{\textbf{bul}} & \multicolumn{1}{c}{\textbf{ces}} & \multicolumn{1}{c}{\textbf{deu}} & \multicolumn{1}{c}{\textbf{ell}} & \multicolumn{1}{c}{\textbf{eng}} & \multicolumn{1}{c}{\textbf{fra}} & \multicolumn{1}{c}{\textbf{hbs}} & \multicolumn{1}{c}{\textbf{hin}} & \multicolumn{1}{c}{\textbf{hun}} & \multicolumn{1}{c}{\textbf{kor}} & \multicolumn{1}{c}{\textbf{msa}} & \multicolumn{1}{c}{\textbf{mya}} & \multicolumn{1}{c}{\textbf{nld}} & \multicolumn{1}{c}{\textbf{pol}} & \multicolumn{1}{c}{\textbf{por}} & \multicolumn{1}{c}{\textbf{ron}} & \multicolumn{1}{c}{\textbf{slk}} & \multicolumn{1}{c}{\textbf{spa}} & \multicolumn{1}{c}{\textbf{tha}} & \multicolumn{1}{c}{\textbf{Avg.}} \\
\midrule
BM25 & Og & 0.75 & 0.71 & 0.70 & 0.63 & 0.61 & 0.63 & 0.74 & 0.46 & 0.61 & 0.49 & 0.58 & 0.75 & 0.31 & 0.56 & 0.56 & 0.77 & 0.70 & 0.78 & 0.73 & 0.31 & 0.62 \\\midrule
\multicolumn{22}{c}{\textit{English TEMs}} \\ \midrule
\texttt{DistilRoBERTa} & En & 0.79 & 0.86 & 0.88 & 0.58 & 0.73 & 0.64 & 0.79 & 0.65 & 0.65 & 0.82 & 0.82 & 0.75 & 0.77 & 0.72 & 0.65 & 0.64 & 0.86 & 0.85 & 0.72 & 0.89 & 0.75 \\
\texttt{MiniLM-L6} & En & 0.84 & 0.89 & 0.85 & 0.64 & 0.80 & 0.69 & 0.82 & 0.70 & 0.75 & 0.87 & 0.84 & 0.78 & 0.79 & 0.76 & 0.70 & 0.70 & 0.86 & 0.84 & 0.77 & \textbf{0.90} & 0.79 \\
\texttt{MiniLM-L12} & En & 0.84 & \textbf{0.90} & 0.86 & 0.64 & 0.80 & 0.70 & 0.82 & 0.72 & 0.77 & 0.86 & 0.83 & 0.78 & 0.80 & 0.73 & 0.72 & 0.71 & 0.86 & 0.86 & 0.78 & 0.89 & 0.79 \\
\texttt{MPNet-Base} & En & 0.80 & 0.87 & \textbf{0.89} & 0.57 & 0.77 & 0.68 & 0.81 & 0.70 & 0.72 & 0.87 & 0.80 & 0.79 & 0.80 & 0.74 & 0.67 & 0.67 & 0.86 & 0.85 & 0.75 & 0.88 & 0.77 \\
\texttt{GTE-Large-En} & En & 0.82 & 0.88 & 0.88 & 0.65 & 0.82 & 0.73 & 0.84 & 0.72 & 0.74 & 0.85 & 0.84 & 0.76 & 0.81 & \textbf{0.78} & 0.71 & 0.69 & 0.87 & 0.86 & 0.79 & 0.89 & 0.80 \\
\texttt{GTR-T5-Large} & En & \textbf{0.86} & 0.86 & 0.88 & \textbf{0.69} & \textbf{0.83} & \textbf{0.77} & \textbf{0.86} & \textbf{0.74} & \textbf{0.79} & \textbf{0.89} & \textbf{0.86} & \textbf{0.82} & \textbf{0.88} & \textbf{0.78} & \textbf{0.74} & \textbf{0.80} & \textbf{0.88} & \textbf{0.87} & \textbf{0.84} & \textbf{0.90} & \textbf{0.83} \\ \midrule
\multicolumn{22}{c}{\textit{Multilingual TEMs}} \\\midrule
\texttt{BGE-M3} & Og & \textbf{0.84} & 0.87 & 0.90 & 0.74 & 0.80 & 0.69 & \textbf{0.87} & 0.67 & \textbf{0.82} & 0.89 & 0.90 & 0.86 & \textbf{0.86} & 0.74 & 0.72 & 0.79 & 0.88 & \textbf{0.89} & 0.84 & \textbf{0.93} & 0.82 \\
\texttt{DistilUSE-Base-Multilingual} & Og & 0.74 & 0.81 & 0.71 & 0.50 & 0.60 & 0.56 & 0.69 & 0.57 & 0.53 & 0.78 & 0.74 & 0.60 & 0.62 & 0.61 & 0.60 & 0.58 & 0.80 & 0.77 & 0.64 & 0.72 & 0.66 \\
\texttt{LaBSE} & Og & 0.77 & 0.84 & 0.81 & 0.48 & 0.70 & 0.44 & 0.72 & 0.57 & 0.56 & 0.82 & 0.77 & 0.67 & 0.77 & 0.61 & 0.57 & 0.66 & 0.78 & 0.74 & 0.64 & 0.79 & 0.69 \\
\texttt{Multilingual E5 Small} & Og & 0.81 & 0.89 & 0.82 & 0.71 & 0.80 & 0.61 & 0.80 & 0.63 & 0.72 & 0.87 & 0.85 & 0.77 & 0.69 & 0.72 & 0.71 & 0.76 & 0.89 & 0.83 & 0.81 & 0.89 & 0.78 \\
\texttt{Multilingual E5 Base} & Og & 0.81 & 0.87 & 0.85 & 0.70 & 0.77 & 0.64 & 0.83 & 0.60 & 0.67 & 0.88 & 0.86 & 0.80 & 0.74 & 0.73 & 0.66 & 0.77 & 0.88 & 0.84 & 0.81 & 0.89 & 0.78 \\
\texttt{Multilingual E5 Large} & Og & \textbf{0.84} & \textbf{0.90} & \textbf{0.92} & \textbf{0.78} & \textbf{0.82} & \textbf{0.75} & 0.86 & \textbf{0.74} & 0.81 & \textbf{0.90} & \textbf{0.91} & \textbf{0.88} & 0.81 & \textbf{0.83} & \textbf{0.77} & \textbf{0.82} & \textbf{0.90} & \textbf{0.89} & \textbf{0.87} & 0.85 & \textbf{0.84} \\
\texttt{MiniLM-L12-Multilingual} & Og & 0.49 & 0.83 & 0.75 & 0.48 & 0.58 & 0.58 & 0.66 & 0.55 & 0.49 & 0.79 & 0.61 & 0.54 & 0.58 & 0.64 & 0.61 & 0.51 & 0.79 & 0.77 & 0.57 & 0.81 & 0.63 \\
\texttt{MPNet-Base-Multilingual} & Og & 0.70 & 0.81 & 0.78 & 0.53 & 0.63 & 0.61 & 0.73 & 0.56 & 0.63 & 0.83 & 0.71 & 0.62 & 0.75 & 0.66 & 0.60 & 0.57 & 0.84 & 0.80 & 0.64 & 0.86 & 0.69 \\
\bottomrule
\end{tabular}
}
\caption{TEM results for retrieving previously fact-checked claims across 20 languages using the S@10 metric. The best scores for each configuration -- English translation (En) or original language (Og) -- are in bold. \texttt{GTR-T-Large} performed best on English translations, while \texttt{Multilingual E5 Large} excelled on multilingual data, surpassing English TEMs.}
\label{tab:retrieval-languages}
\end{table*}

\subsection{Criteria-based Retrieval}

\begin{figure}[t]
    \centering
    \includegraphics[width=\columnwidth]{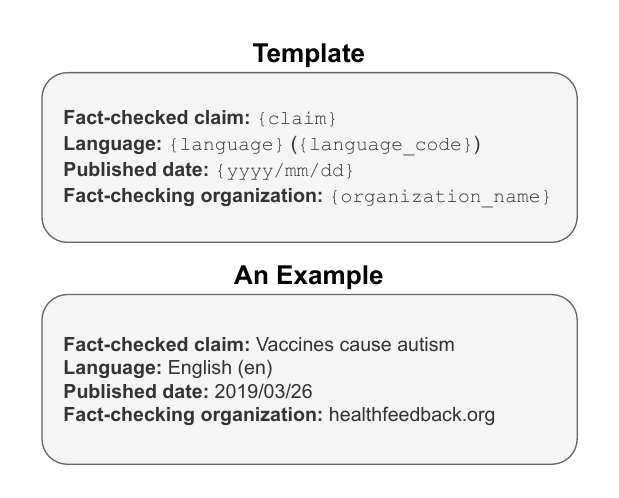}
    \caption{Template used to structure fact-checks for filtered retrieval, along with an example illustrating its format, including the fact-checked claim, language, publication date and fact-checking organization.}
    \label{fig:template}
\end{figure}

Figure~\ref{fig:template} illustrates the template used for filtered retrieval experiments. Each fact-check is structured using this template, which includes the fact-checked claim, the language of the fact-checking article, the publication date, and the fact-checking organization. This structure representation is then embedded using the selected TEM.

To retrieve relevant fact-checks based on the instruction in natural language, we test different retrieval conditions, such as filtering by language or by a specific named entity. Once we obtain a list of fact-checks with a similarity score above 0.8, we perform a second retrieval step based on the content of a social media post. In this step, each fact-check is represented only by the fact-checked claim without any metadata, and is embedded using a specific TEM to facilitate retrieval.

\section{Summarization Experiments}

\begin{figure}
    \centering
    \includegraphics[width=\columnwidth]{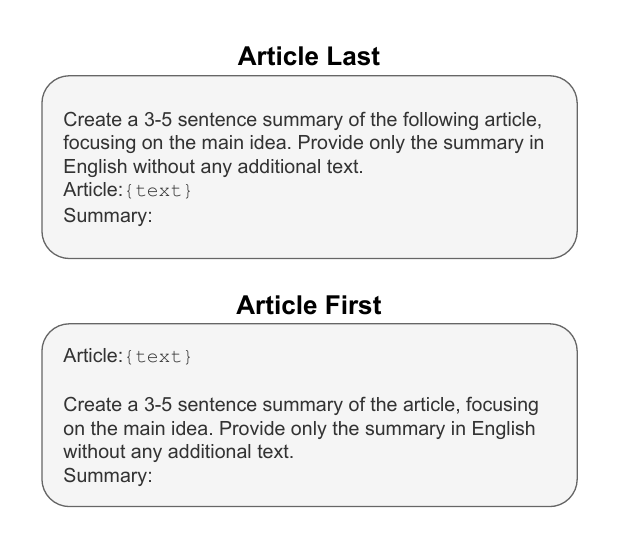}
    \caption{Prompts used for the experiments with summarization.}
    \label{fig:summary_prompts}
\end{figure}

Figure~\ref{fig:summary_prompts} illustrate the final prompt formats used in our summarization experiments. We present both the \textit{Article last} and \textit{Article first} variants.

Table~\ref{tab:summary-bertscore} presents the results of the summarization experiments using various open-source and closed-source LLMs across 23 languages, evaluated with the BERTScore metric. For each LLM, we report performance in two settings: when the article is provided before the instruction (\textit{Article first}) and when the article is provided after the instruction (\textit{Article last}).

Similarly, Table~\ref{tab:summary-rouge} summarizes the results based on the ROUGE-L metric.

In addition to evaluating the two settings, we also examined the impact of different quantization variants on LLM performance. Specifically, we compared non-quantized models with versions quantized to 4-bit and 8-bit precision. For these experiments, we selected \texttt{Llama} models, focusing on \texttt{Llama 3.1 70B} and \texttt{Llama3.2} in 1B and 3B variants. The BERTscore results across 23 languages are presented in Table~\ref{tab:summary-quantization}, while Table~\ref{tab:rouge-quant} reports ROUGE-L scores.

\begin{table*}
\resizebox{\textwidth}{!}{%
\begin{tabular}{lllrrrrrrrrrrrrrrrrrrrrrrrr}
\toprule
\multicolumn{1}{c}{\textbf{Model}} & \multicolumn{1}{c}{\textbf{Version}} & \multicolumn{1}{c}{\textbf{Quant.}} & \multicolumn{1}{c}{\textbf{bg}} & \multicolumn{1}{c}{\textbf{bn}} & \multicolumn{1}{c}{\textbf{ca}} & \multicolumn{1}{c}{\textbf{cs}} & \multicolumn{1}{c}{\textbf{de}} & \multicolumn{1}{c}{\textbf{el}} & \multicolumn{1}{c}{\textbf{en}} & \multicolumn{1}{c}{\textbf{es}} & \multicolumn{1}{c}{\textbf{fi}} & \multicolumn{1}{c}{\textbf{fr}} & \multicolumn{1}{c}{\textbf{hi}} & \multicolumn{1}{c}{\textbf{hu}} & \multicolumn{1}{c}{\textbf{id}} & \multicolumn{1}{c}{\textbf{ko}} & \multicolumn{1}{c}{\textbf{ms}} & \multicolumn{1}{c}{\textbf{my}} & \multicolumn{1}{c}{\textbf{nl}} & \multicolumn{1}{c}{\textbf{pl}} & \multicolumn{1}{c}{\textbf{pt}} & \multicolumn{1}{c}{\textbf{ro}} & \multicolumn{1}{c}{\textbf{sk}} & \multicolumn{1}{c}{\textbf{sr}} & \multicolumn{1}{c}{\textbf{th}} & \multicolumn{1}{c}{\textbf{Avg.}} \\ \midrule
\multicolumn{27}{c}{\textit{Open-Source LLMs}} \\ \midrule
\multirow{2}{*}{C4AI Command R+} & Article first & 4bit & \textbf{0.75} & 0.76 & \textbf{0.74} & 0.74 & \textbf{0.75} & \textbf{0.76} & 0.76 & 0.73 & \textbf{0.75} & \textbf{0.74} & 0.76 & \textbf{0.75} & \textbf{0.77} & \textbf{0.77} & \textbf{0.76} & \textbf{0.74} & \textbf{0.75} & \textbf{0.75} & \textbf{0.74} & \textbf{0.76} & \textbf{0.75} & \textbf{0.75} & \textbf{0.75} & \textbf{0.75} \\
 & Article last & 4bit & 0.70 & 0.71 & 0.72 & 0.69 & 0.71 & 0.67 & 0.76 & 0.70 & 0.71 & 0.69 & 0.71 & 0.68 & 0.72 & 0.75 & 0.74 & 0.66 & 0.73 & 0.70 & 0.72 & 0.69 & 0.73 & 0.73 & 0.64 & 0.71 \\ \midrule
\multirow{2}{*}{Llama3.1 70B Instruct} & Article first & 4bit & \textbf{0.75} & \textbf{0.77} & \textbf{0.74} & \textbf{0.74} & \textbf{0.75} & \textbf{0.76} & \textbf{0.76} & 0.73 & \textbf{0.75} & \textbf{0.74} & 0.76 & \textbf{0.75} & \textbf{0.77} & \textbf{0.77} & \textbf{0.76} & 0.72 & \textbf{0.75} & \textbf{0.75} & \textbf{0.74} & \textbf{0.76} & \textbf{0.75} & \textbf{0.75} & \textbf{0.75} & \textbf{0.75} \\
 & Article last & 4bit & \textbf{0.75} & \textbf{0.77} & \textbf{0.74} & \textbf{0.74} & \textbf{0.75} & \textbf{0.76} & \textbf{0.76} & \textbf{0.74} & 0.74 & \textbf{0.74} & \textbf{0.77} & 0.69 & \textbf{0.77} & \textbf{0.77} & \textbf{0.76} & 0.72 & \textbf{0.75} & \textbf{0.75} & \textbf{0.74} & 0.75 & \textbf{0.75} & \textbf{0.75} & \textbf{0.75} & \textbf{0.75} \\ \midrule
\multirow{2}{*}{Llama3.3 70B Instruct} & Article first & 4bit & 0.74 & 0.76 & 0.73 & \textbf{0.74} & 0.74 & 0.75 & 0.75 & 0.73 & \textbf{0.75} & 0.73 & 0.76 & \textbf{0.75} & 0.76 & 0.76 & 0.75 & 0.70 & \textbf{0.75} & 0.74 & \textbf{0.74} & \textbf{0.76} & 0.74 & 0.74 & \textbf{0.75} & 0.74 \\
 & Article last & 4bit & \textbf{0.75} & 0.76 & 0.73 & \textbf{0.74} & 0.74 & 0.75 & 0.75 & 0.73 & \textbf{0.75} & 0.73 & 0.76 & 0.74 & 0.76 & 0.76 & 0.75 & 0.70 & \textbf{0.75} & 0.74 & 0.73 & 0.75 & 0.74 & 0.74 & \textbf{0.75} & 0.74 \\ \midrule
\multirow{2}{*}{Mistral Large} & Article first & 4bit & \textbf{0.75} & \textbf{0.77} & 0.73 & \textbf{0.74} & \textbf{0.75} & \textbf{0.76} & \textbf{0.76} & 0.73 & \textbf{0.75} & \textbf{0.74} & 0.75 & \textbf{0.75} & \textbf{0.77} & \textbf{0.77} & \textbf{0.76} & \textbf{0.74} & \textbf{0.75} & \textbf{0.75} & \textbf{0.74} & 0.75 & 0.74 & 0.74 & \textbf{0.75} & \textbf{0.75} \\
 & Article last & 4bit & \textbf{0.75} & 0.76 & \textbf{0.74} & \textbf{0.74} & \textbf{0.75} & 0.75 & \textbf{0.76} & 0.73 & \textbf{0.75} & \textbf{0.74} & 0.75 & 0.74 & \textbf{0.77} & \textbf{0.77} & \textbf{0.76} & \textbf{0.74} & \textbf{0.75} & 0.74 & \textbf{0.74} & 0.75 & 0.74 & 0.74 & \textbf{0.75} & \textbf{0.75} \\ \midrule
\multirow{2}{*}{Qwen2 72B Instruct} & Article first & 4bit & 0.74 & 0.75 & 0.73 & 0.73 & 0.74 & 0.75 & 0.75 & 0.72 & 0.74 & 0.73 & 0.75 & 0.73 & 0.76 & 0.76 & 0.74 & \textbf{0.74} & 0.74 & 0.74 & 0.73 & 0.74 & 0.74 & 0.74 & 0.74 & 0.74 \\
 & Article last & 4bit & 0.74 & 0.76 & 0.73 & 0.73 & 0.74 & 0.74 & 0.75 & 0.72 & 0.74 & 0.73 & 0.75 & 0.74 & 0.76 & 0.76 & 0.74 & 0.73 & 0.74 & 0.74 & 0.73 & 0.75 & 0.74 & 0.73 & 0.74 & 0.74 \\ \midrule
\multirow{2}{*}{Qwen2.5 0.5B Instruct} & Article first & - & 0.70 & 0.68 & 0.71 & 0.70 & 0.72 & 0.69 & 0.74 & 0.71 & 0.68 & 0.71 & 0.69 & 0.69 & 0.73 & 0.72 & 0.71 & 0.68 & 0.71 & 0.71 & 0.71 & 0.71 & 0.70 & 0.69 & 0.72 & 0.70 \\
 & Article last & - & 0.70 & 0.68 & 0.70 & 0.70 & 0.71 & 0.68 & 0.73 & 0.70 & 0.68 & 0.70 & 0.68 & 0.69 & 0.73 & 0.72 & 0.71 & 0.67 & 0.70 & 0.70 & 0.70 & 0.71 & 0.70 & 0.69 & 0.72 & 0.70 \\ \midrule
\multirow{2}{*}{Qwen2.5 1.5B Instruct} & Article first & - & 0.73 & 0.73 & 0.73 & 0.72 & 0.73 & 0.72 & 0.75 & 0.73 & 0.71 & 0.73 & 0.73 & 0.72 & 0.76 & 0.74 & 0.74 & 0.70 & 0.74 & 0.73 & 0.73 & 0.74 & 0.73 & 0.72 & 0.74 & 0.73 \\
 & Article last & - & 0.73 & 0.73 & 0.72 & 0.72 & 0.73 & 0.72 & 0.75 & 0.72 & 0.72 & 0.72 & 0.73 & 0.72 & 0.75 & 0.75 & 0.74 & 0.69 & 0.73 & 0.73 & 0.72 & 0.73 & 0.72 & 0.72 & 0.74 & 0.73 \\ \midrule
\multirow{2}{*}{Qwen2.5 3B Instruct} & Article first & - & 0.74 & 0.75 & 0.73 & 0.73 & 0.74 & 0.74 & \textbf{0.76} & 0.73 & 0.73 & 0.73 & 0.75 & 0.73 & 0.76 & 0.75 & 0.75 & 0.71 & 0.74 & 0.74 & \textbf{0.74} & 0.75 & 0.74 & 0.74 & 0.74 & 0.74 \\
 & Article last & - & 0.71 & 0.71 & 0.71 & 0.69 & 0.71 & 0.71 & 0.74 & 0.71 & 0.71 & 0.72 & 0.72 & 0.71 & 0.72 & 0.74 & 0.72 & 0.68 & 0.71 & 0.69 & 0.71 & 0.72 & 0.69 & 0.68 & 0.73 & 0.71 \\ \midrule
\multirow{2}{*}{Qwen2.5 7B Instruct} & Article first & - & 0.73 & 0.75 & 0.73 & 0.73 & 0.74 & 0.74 & 0.75 & 0.72 & 0.73 & 0.73 & 0.74 & 0.73 & 0.76 & 0.75 & 0.75 & 0.70 & 0.73 & 0.73 & 0.73 & 0.74 & 0.73 & 0.73 & 0.74 & 0.74 \\
 & Article last & - & 0.74 & 0.75 & 0.73 & 0.73 & 0.74 & 0.74 & \textbf{0.76} & 0.73 & 0.74 & 0.73 & 0.74 & 0.73 & 0.76 & 0.75 & 0.75 & 0.72 & 0.74 & 0.74 & \textbf{0.74} & 0.75 & 0.74 & 0.74 & \textbf{0.75} & 0.74 \\ \midrule
\multirow{2}{*}{Qwen2.5 72B Instruct} & Article first & 4bit & \textbf{0.75} & \textbf{0.77} & \textbf{0.74} & \textbf{0.74} & 0.74 & 0.75 & \textbf{0.76} & 0.73 & \textbf{0.75} & \textbf{0.74} & 0.76 & 0.74 & \textbf{0.77} & 0.76 & \textbf{0.76} & \textbf{0.74} & \textbf{0.75} & \textbf{0.75} & \textbf{0.74} & 0.75 & \textbf{0.75} & \textbf{0.75} & \textbf{0.75} & \textbf{0.75} \\
 & Article last & 4bit & \textbf{0.75} & 0.76 & \textbf{0.74} & \textbf{0.74} & 0.74 & 0.75 & 0.75 & 0.73 & \textbf{0.75} & \textbf{0.74} & 0.76 & \textbf{0.75} & \textbf{0.77} & 0.76 & \textbf{0.76} & \textbf{0.74} & \textbf{0.75} & 0.74 & \textbf{0.74} & \textbf{0.76} & 0.74 & \textbf{0.75} & \textbf{0.75} & \textbf{0.75} \\ \midrule
 \multirow{2}{*}{Gemma3 27B} & Article first & 4bit & 0.72 & 0.75 & 0.72 & 0.72 & 0.73 & 0.74 & 0.74 & 0.72 & 0.73 & 0.73 & 0.74 & 0.73 & 0.75 & 0.75 & 0.74 & 0.73 & 0.73 & 0.72 & 0.72 & 0.73 & 0.73 & 0.72 & 0.73 & 0.73 \\
 & Article last & 4bit & 0.72 & 0.74 & 0.72 & 0.73 & 0.73 & 0.74 & 0.74 & 0.72 & 0.73 & 0.72 & 0.74 & 0.73 & 0.75 & 0.74 & 0.74 & 0.72 & 0.73 & 0.73 & 0.72 & 0.73 & 0.73 & 0.73 & 0.73 & 0.73\\
\midrule
\multicolumn{27}{c}{\textit{Closed-Source LLMs}} \\ \midrule
\multirow{2}{*}{Claude 3.5 Sonnet} & Article first & - &  0.74 & 0.76 & 0.73 & \textbf{0.74} & 0.73 & 0.74 & 0.75 & 0.72 & 0.74 & 0.73 & 0.75 & 0.74 & 0.76 & 0.75 & 0.75 & 0.73 & 0.74 & 0.74 & 0.73 & 0.74 & 0.74 & 0.73 & \textbf{0.75} & 0.74 \\
 & Article last & - & 0.74 & 0.76 & 0.73 & \textbf{0.74} & 0.74 & 0.75 & \textbf{0.76} & 0.73 & 0.74 & \textbf{0.74} & 0.75 & 0.74 & 0.76 & 0.76 & 0.75 & \textbf{0.74} & 0.74 & 0.74 & \textbf{0.74} & 0.74 & 0.74 & 0.74 & \textbf{0.75} & 0.74 \\ \midrule
\multirow{2}{*}{GPT-4o} & Article first & - & 0.74 & 0.76 & 0.73 & \textbf{0.74} & 0.74 & 0.75 & 0.75 & 0.72 & 0.74 & 0.73 & 0.75 & 0.74 & 0.76 & 0.76 & 0.75 & \textbf{0.74} & 0.74 & 0.74 & \textbf{0.74} & 0.74 & 0.74 & 0.74 & \textbf{0.75} & 0.74 \\
 & Article last & - & 0.74 & 0.76 & 0.73 & 0.73 & 0.74 & 0.75 & 0.75 & 0.72 & 0.74 & 0.73 & 0.75 & 0.74 & 0.76 & 0.76 & 0.75 & \textbf{0.74} & 0.74 & 0.74 & 0.73 & 0.74 & 0.74 & 0.74 & \textbf{0.75} & 0.74 \\
 \bottomrule
\end{tabular}
}
\caption{BERTScore evaluation of summarization performance across 23 languages for various LLMs in two settings: \textit{Article first} and \textit{Article last}. The best results for each language are in bold.}
\label{tab:summary-bertscore}
\end{table*}

\begin{table*}[]
\resizebox{\textwidth}{!}{%
\begin{tabular}{lccrrrrrrrrrrrrrrrrrrrrrrrr}
\toprule
\multicolumn{1}{c}{Model} & \textbf{Version} & \textbf{Quant.} & \multicolumn{1}{c}{\textbf{bg}} & \multicolumn{1}{c}{\textbf{bn}} & \multicolumn{1}{c}{\textbf{ca}} & \multicolumn{1}{c}{\textbf{cs}} & \multicolumn{1}{c}{\textbf{de}} & \multicolumn{1}{c}{\textbf{el}} & \multicolumn{1}{c}{\textbf{en}} & \multicolumn{1}{c}{\textbf{es}} & \multicolumn{1}{c}{\textbf{fi}} & \multicolumn{1}{c}{\textbf{fr}} & \multicolumn{1}{c}{\textbf{hi}} & \multicolumn{1}{c}{\textbf{hu}} & \multicolumn{1}{c}{\textbf{id}} & \multicolumn{1}{c}{\textbf{ko}} & \multicolumn{1}{c}{\textbf{ms}} & \multicolumn{1}{c}{\textbf{my}} & \multicolumn{1}{c}{\textbf{nl}} & \multicolumn{1}{c}{\textbf{pl}} & \multicolumn{1}{c}{\textbf{pt}} & \multicolumn{1}{c}{\textbf{ro}} & \multicolumn{1}{c}{\textbf{sk}} & \multicolumn{1}{c}{\textbf{sr}} & \multicolumn{1}{c}{\textbf{th}} & \multicolumn{1}{c}{\textbf{Avg.}} \\ \midrule
\multicolumn{27}{c}{\textit{Open-Source LLMs}} \\ \midrule
\multirow{2}{*}{C4AI Command R+} & Article first & 4bit & \textbf{0.32} & 0.34 & 0.28 & 0.28 & 0.30 & \textbf{0.32} & \textbf{0.33} & 0.27 & 0.30 & \textbf{0.28} & 0.34 & 0.30 & \textbf{0.37} & 0.34 & \textbf{0.35} & \textbf{0.28} & 0.31 & \textbf{0.30} & 0.30 & 0.32 & 0.29 & \textbf{0.30} & 0.32 & \textbf{0.31} \\
 & Article last & 4bit & 0.12 & 0.19 & 0.18 & 0.12 & 0.12 & 0.08 & 0.32 & 0.08 & 0.19 & 0.05 & 0.16 & 0.10 & 0.11 & 0.28 & 0.24 & 0.15 & 0.23 & 0.14 & 0.11 & 0.06 & 0.25 & 0.26 & 0.09 & 0.16 \\ \midrule
\multirow{2}{*}{Llama3.1 70B Instruct} & Article first & 4bit & 0.31 & 0.34 & \textbf{0.29} & \textbf{0.29} & \textbf{0.31} & \textbf{0.32} & \textbf{0.33} & \textbf{0.29} & \textbf{0.31} & \textbf{0.28} & 0.34 & \textbf{0.31} & \textbf{0.37} & 0.34 & 0.34 & 0.24 & \textbf{0.32} & \textbf{0.30} & \textbf{0.31} & \textbf{0.33} & 0.28 & \textbf{0.30} & 0.33 & \textbf{0.31} \\
 & Article last & 4bit & 0.30 & 0.32 & 0.28 & 0.28 & \textbf{0.31} & 0.30 & \textbf{0.33} & 0.28 & 0.28 & 0.27 & 0.33 & 0.13 & 0.34 & 0.34 & 0.32 & 0.24 & 0.31 & \textbf{0.30} & 0.28 & 0.27 & 0.28 & 0.29 & 0.32 & 0.29 \\ \midrule
\multirow{2}{*}{Llama3.3 70B Instruct} & Article first & 4bit & 0.30 & \textbf{0.35} & \textbf{0.29} & \textbf{0.29} & \textbf{0.31} & \textbf{0.32} & 0.32 & 0.28 & \textbf{0.31} & 0.27 & \textbf{0.35} & 0.30 & \textbf{0.37} & 0.34 & 0.34 & 0.22 & 0.31 & \textbf{0.30} & 0.30 & \textbf{0.33} & 0.29 & 0.29 & \textbf{0.34} & \textbf{0.31} \\
 & Article last & 4bit & 0.31 & 0.34 & \textbf{0.29} & 0.28 & \textbf{0.31} & 0.31 & 0.32 & \textbf{0.29} & \textbf{0.31} & 0.27 & \textbf{0.35} & 0.26 & 0.35 & \textbf{0.35} & 0.34 & 0.22 & 0.31 & 0.29 & 0.30 & 0.31 & 0.29 & 0.29 & \textbf{0.34} & \textbf{0.31} \\ \midrule
\multirow{2}{*}{Mistral Large} & Article first & 4bit & 0.31 & 0.33 & 0.27 & 0.28 & 0.30 & \textbf{0.32} & \textbf{0.33} & 0.27 & 0.30 & 0.27 & 0.32 & 0.30 & 0.35 & 0.34 & 0.33 & 0.27 & 0.31 & 0.29 & 0.30 & 0.30 & 0.28 & 0.29 & 0.32 & 0.30 \\
 & Article last & 4bit & 0.30 & 0.33 & 0.28 & \textbf{0.29} & \textbf{0.31} & \textbf{0.32} & 0.31 & 0.27 & \textbf{0.31} & 0.27 & 0.33 & 0.29 & 0.36 & 0.34 & 0.33 & \textbf{0.28} & 0.31 & 0.29 & 0.29 & 0.31 & \textbf{0.30} & 0.29 & 0.33 & \textbf{0.31} \\ \midrule
\multirow{2}{*}{Qwen2 72B Instruct} & Article first & 4bit & 0.28 & 0.31 & 0.25 & 0.26 & 0.27 & 0.29 & 0.29 & 0.25 & 0.28 & 0.24 & 0.30 & 0.27 & 0.32 & 0.31 & 0.30 & 0.26 & 0.28 & 0.27 & 0.26 & 0.28 & 0.27 & 0.27 & 0.29 & 0.28 \\
 & Article last & 4bit & 0.28 & 0.32 & 0.25 & 0.26 & 0.28 & 0.29 & 0.30 & 0.24 & 0.28 & 0.24 & 0.29 & 0.28 & 0.32 & 0.31 & 0.29 & 0.26 & 0.28 & 0.27 & 0.27 & 0.28 & 0.26 & 0.27 & 0.30 & 0.28 \\ \midrule
\multirow{2}{*}{Qwen2.5 0.5B Instruct} & Article first & - & 0.19 & 0.16 & 0.19 & 0.18 & 0.24 & 0.16 & 0.25 & 0.21 & 0.15 & 0.19 & 0.17 & 0.16 & 0.25 & 0.23 & 0.21 & 0.15 & 0.21 & 0.19 & 0.20 & 0.19 & 0.17 & 0.16 & 0.24 & 0.19 \\
 & Article last & - & 0.20 & 0.16 & 0.18 & 0.19 & 0.21 & 0.16 & 0.25 & 0.18 & 0.15 & 0.19 & 0.18 & 0.16 & 0.24 & 0.24 & 0.22 & 0.15 & 0.18 & 0.19 & 0.17 & 0.19 & 0.17 & 0.17 & 0.24 & 0.19 \\ \midrule
\multirow{2}{*}{Qwen2.5 1.5B Instruct} & Article first & - & 0.25 & 0.24 & 0.24 & 0.22 & 0.25 & 0.23 & 0.26 & 0.25 & 0.21 & 0.24 & 0.26 & 0.21 & 0.31 & 0.26 & 0.28 & 0.17 & 0.27 & 0.25 & 0.25 & 0.27 & 0.23 & 0.23 & 0.29 & 0.25 \\
 & Article last & - & 0.26 & 0.24 & 0.22 & 0.21 & 0.24 & 0.23 & 0.26 & 0.25 & 0.23 & 0.23 & 0.25 & 0.21 & 0.29 & 0.28 & 0.27 & 0.17 & 0.25 & 0.24 & 0.23 & 0.26 & 0.24 & 0.23 & 0.29 & 0.24 \\ \midrule
\multirow{2}{*}{Qwen2.5 3B Instruct} & Article first & - & 0.27 & 0.28 & 0.25 & 0.24 & 0.27 & 0.27 & 0.30 & 0.25 & 0.26 & 0.24 & 0.28 & 0.25 & 0.31 & 0.30 & 0.28 & 0.21 & 0.27 & 0.26 & 0.26 & 0.28 & 0.25 & 0.26 & 0.30 & 0.27 \\
 & Article last & - & 0.24 & 0.25 & 0.23 & 0.19 & 0.25 & 0.23 & 0.28 & 0.23 & 0.21 & 0.23 & 0.24 & 0.22 & 0.27 & 0.29 & 0.26 & 0.16 & 0.25 & 0.20 & 0.23 & 0.25 & 0.18 & 0.17 & 0.29 & 0.23 \\ \midrule
\multirow{2}{*}{Qwen2.5 7B Instruct} & Article first & - & 0.25 & 0.28 & 0.23 & 0.24 & 0.26 & 0.26 & 0.26 & 0.23 & 0.24 & 0.23 & 0.27 & 0.25 & 0.30 & 0.26 & 0.27 & 0.17 & 0.25 & 0.25 & 0.24 & 0.26 & 0.25 & 0.24 & 0.27 & 0.25 \\
 & Article last & - & 0.28 & 0.30 & 0.25 & 0.26 & 0.28 & 0.28 & 0.29 & 0.25 & 0.27 & 0.24 & 0.29 & 0.25 & 0.32 & 0.31 & 0.30 & 0.23 & 0.28 & 0.26 & 0.27 & 0.29 & 0.27 & 0.27 & 0.31 & 0.28 \\ \midrule
\multirow{2}{*}{Qwen2.5 72B Instruct} & Article first & 4bit & 0.29 & 0.32 & 0.27 & 0.28 & 0.29 & 0.29 & 0.30 & 0.26 & 0.30 & 0.25 & 0.31 & 0.29 & 0.34 & 0.32 & 0.32 & 0.26 & 0.29 & 0.29 & 0.28 & 0.29 & 0.28 & 0.28 & 0.31 & 0.29 \\
 & Article last & 4bit & 0.29 & 0.32 & 0.26 & 0.27 & 0.28 & 0.30 & 0.30 & 0.25 & 0.30 & 0.25 & 0.31 & 0.29 & 0.34 & 0.32 & 0.31 & 0.25 & 0.29 & 0.26 & 0.28 & 0.29 & 0.28 & 0.29 & 0.31 & 0.29 \\ \midrule
\multirow{2}{*}{Gemma3 27B} & Article first & 4bit &0.26 & 0.30 & 0.24 & 0.25 & 0.26 & 0.26 & 0.29 & 0.24 & 0.26 & 0.25 & 0.30 & 0.26 & 0.32 & 0.29 & 0.31 & 0.26 & 0.27 & 0.25 & 0.25 & 0.26 & 0.25 & 0.25 & 0.30 & 0.27 \\
& Article last & 4bit & 0.26 & 0.30 & 0.24 & 0.25 & 0.27 & 0.27 & 0.28 & 0.24 & 0.26 & 0.24 & 0.30 & 0.27 & 0.31 & 0.29 & 0.30 & 0.25 & 0.26 & 0.25 & 0.25 & 0.27 & 0.25 & 0.26 & 0.30 & 0.27 \\ \midrule
\multicolumn{27}{c}{\textit{Closed-Source LLMs}} \\ \midrule
\multirow{2}{*}{Claude 3.5 Sonnet} & Article first & - & 0.30 & 0.34 & 0.26 & 0.28 & 0.29 & 0.29 & 0.30 & 0.26 & 0.29 & 0.27 & 0.33 & 0.30 & 0.33 & 0.32 & 0.31 & 0.27 & 0.29 & 0.28 & 0.28 & 0.30 & 0.28 & 0.27 & 0.33 & 0.29 \\
 & Article last & - & 0.29 & 0.33 & 0.27 & \textbf{0.29} & 0.30 & 0.30 & 0.32 & 0.27 & 0.29 & 0.27 & 0.33 & 0.28 & 0.34 & 0.34 & 0.31 & \textbf{0.28} & 0.29 & 0.29 & 0.29 & 0.29 & 0.28 & 0.28 & 0.33 & 0.30 \\ \midrule
\multirow{2}{*}{GPT 4o} & Article first & - & 0.29 & 0.32 & 0.27 & 0.28 & 0.29 & 0.30 & 0.30 & 0.26 & 0.29 & 0.27 & 0.33 & 0.28 & 0.33 & 0.33 & 0.31 & 0.27 & 0.28 & 0.28 & 0.28 & 0.29 & 0.28 & 0.27 & 0.31 & 0.29 \\
 & Article last & - & 0.29 & 0.33 & 0.26 & 0.27 & 0.29 & 0.29 & 0.30 & 0.26 & 0.29 & 0.25 & 0.32 & 0.29 & 0.32 & 0.33 & 0.31 & 0.26 & 0.28 & 0.27 & 0.27 & 0.29 & 0.27 & 0.28 & 0.31 & 0.29 \\ \bottomrule
\end{tabular}
}
\caption{ROUGE-L evaluation of summarization performance across 23 languages for various LLMs in two settings: \textit{Article first} and \textit{Article last}. The best results for each language are in bold.}
\label{tab:summary-rouge}
\end{table*}

\begin{table*}[]
\resizebox{\textwidth}{!}{%
\begin{tabular}{llcrrrrrrrrrrrrrrrrrrrrrrrr}
\toprule
\multicolumn{1}{c}{\textbf{Model}} & \multicolumn{1}{c}{\textbf{Version}} & \textbf{Quant.} & \multicolumn{1}{c}{\textbf{bg}} & \multicolumn{1}{c}{\textbf{bn}} & \multicolumn{1}{c}{\textbf{ca}} & \multicolumn{1}{c}{\textbf{cs}} & \multicolumn{1}{c}{\textbf{de}} & \multicolumn{1}{c}{\textbf{el}} & \multicolumn{1}{c}{\textbf{en}} & \multicolumn{1}{c}{\textbf{es}} & \multicolumn{1}{c}{\textbf{fi}} & \multicolumn{1}{c}{\textbf{fr}} & \multicolumn{1}{c}{\textbf{hi}} & \multicolumn{1}{c}{\textbf{hu}} & \multicolumn{1}{c}{\textbf{id}} & \multicolumn{1}{c}{\textbf{ko}} & \multicolumn{1}{c}{\textbf{ms}} & \multicolumn{1}{c}{\textbf{my}} & \multicolumn{1}{c}{\textbf{nl}} & \multicolumn{1}{c}{\textbf{pl}} & \multicolumn{1}{c}{\textbf{pt}} & \multicolumn{1}{c}{\textbf{ro}} & \multicolumn{1}{c}{\textbf{sk}} & \multicolumn{1}{c}{\textbf{sr}} & \multicolumn{1}{c}{\textbf{th}} & \multicolumn{1}{c}{\textbf{Avg.}} \\ \midrule
\multirow{6}{*}{Llama3.2 1B Instruct} & \multirow{3}{*}{Article first} & 4bit & 0.70 & 0.67 & 0.71 & 0.69 & 0.71 & 0.70 & 0.74 & 0.70 & 0.66 & 0.69 & 0.64 & 0.64 & 0.72 & 0.72 & 0.70 & 0.65 & 0.72 & 0.68 & 0.70 & 0.70 & 0.69 & 0.68 & 0.60 & 0.69 \\
 &  & 8bit & \textbf{0.72} & \textbf{0.72} & 0.71 & \textbf{0.71} & \textbf{0.73} & \textbf{0.73} & \textbf{0.74} & \textbf{0.71} & \textbf{0.70} & \textbf{0.72} & \textbf{0.72} & \textbf{0.67} & \textbf{0.74} & \textbf{0.73} & \textbf{0.73} & 0.66 & \textbf{0.73} & \textbf{0.71} & \textbf{0.72} & 0.72 & \textbf{0.72} & \textbf{0.71} & \textbf{0.72} & \textbf{0.72} \\
 &  & - & \textbf{0.72} & \textbf{0.72} & \textbf{0.72} & \textbf{0.71} & \textbf{0.73} & \textbf{0.73} & \textbf{0.74} & \textbf{0.71} & \textbf{0.70} & 0.71 & \textbf{0.72} & 0.66 & \textbf{0.74} & \textbf{0.73} & \textbf{0.73} & \textbf{0.67} & 0.72 & \textbf{0.71} & 0.71 & \textbf{0.73} & 0.71 & \textbf{0.71} & \textbf{0.72} & \textbf{0.72} \\ \cline{2-27}
 & \multirow{3}{*}{Article last} & 4bit & 0.61 & 0.60 & 0.66 & 0.64 & 0.67 & 0.62 & 0.74 & 0.68 & 0.61 & 0.67 & 0.63 & 0.63 & 0.69 & 0.63 & 0.67 & 0.52 & 0.67 & 0.65 & 0.67 & 0.65 & 0.61 & 0.62 & 0.59 & 0.64 \\
 &  & 8bit & 0.64 & 0.63 & 0.68 & 0.67 & 0.70 & 0.64 & 0.75 & 0.69 & 0.64 & 0.70 & 0.64 & 0.65 & 0.71 & 0.69 & 0.69 & 0.53 & 0.69 & 0.67 & 0.69 & 0.67 & 0.65 & 0.65 & 0.61 & 0.66 \\
 &  & - & 0.64 & 0.63 & 0.69 & 0.67 & 0.70 & 0.64 & 0.75 & 0.70 & 0.64 & 0.70 & 0.64 & 0.64 & 0.71 & 0.69 & 0.69 & 0.53 & 0.69 & 0.67 & 0.69 & 0.67 & 0.65 & 0.66 & 0.61 & 0.66 \\ \midrule
\multirow{6}{*}{Llama3.2 3B Instruct} & \multirow{3}{*}{Article first} & 4bit & \textbf{0.74} & 0.75 & \textbf{0.73} & \textbf{0.73} & \textbf{0.74} & 0.74 & \textbf{0.76} & \textbf{0.73} & 0.72 & \textbf{0.73} & \textbf{0.75} & 0.69 & \textbf{0.76} & 0.75 & 0.74 & 0.69 & \textbf{0.74} & \textbf{0.74} & \textbf{0.72} & \textbf{0.75} & \textbf{0.73} & \textbf{0.73} & \textbf{0.74} & 0.73 \\
 &  & 8bit & \textbf{0.74} & 0.75 & \textbf{0.73} & \textbf{0.73} & \textbf{0.74} & 0.74 & 0.75 & 0.72 & \textbf{0.74} & \textbf{0.73} & \textbf{0.75} & \textbf{0.70} & 0.75 & 0.75 & 0.74 & \textbf{0.70} & \textbf{0.74} & 0.73 & \textbf{0.72} & \textbf{0.75} & \textbf{0.73} & \textbf{0.73} & \textbf{0.74} & \textbf{0.74} \\
 &  & - & 0.73 & 0.75 & \textbf{0.73} & \textbf{0.73} & \textbf{0.74} & \textbf{0.75} & \textbf{0.76} & 0.72 & \textbf{0.74} & \textbf{0.73} & \textbf{0.75} & 0.69 & 0.75 & \textbf{0.76} & 0.74 & \textbf{0.70} & \textbf{0.74} & \textbf{0.74} & \textbf{0.72} & \textbf{0.75} & \textbf{0.73} & \textbf{0.73} & \textbf{0.74} & \textbf{0.74} \\ \cline{2-27}
 & \multirow{3}{*}{Article last} & 4bit & 0.73 & 0.75 & 0.72 & 0.72 & 0.73 & 0.73 & 0.76 & 0.72 & 0.70 & \textbf{0.73} & \textbf{0.75} & 0.66 & \textbf{0.76} & \textbf{0.76} & \textbf{0.75} & 0.68 & 0.73 & 0.72 & 0.71 & 0.73 & 0.72 & 0.72 & \textbf{0.74} & 0.73 \\
 &  & 8bit & 0.70 & \textbf{0.76} & 0.71 & 0.69 & 0.72 & 0.66 & 0.75 & 0.72 & 0.67 & \textbf{0.73} & 0.73 & 0.66 & 0.74 & \textbf{0.76} & 0.73 & 0.69 & 0.72 & 0.71 & 0.71 & 0.69 & 0.68 & 0.69 & 0.73 & 0.71 \\
 &  & - & 0.69 & \textbf{0.76} & 0.70 & 0.69 & 0.72 & 0.67 & 0.75 & 0.72 & 0.66 & \textbf{0.73} & 0.73 & 0.66 & 0.74 & \textbf{0.76} & 0.73 & 0.70 & 0.72 & 0.71 & 0.71 & 0.69 & 0.68 & 0.69 & \textbf{0.74} & 0.71 \\ \midrule
 \multirow{2}{*}{Llama3.1 70B Instruct} & \multirow{2}{*}{Article first} & 4bit & \textbf{0.75} & \textbf{0.77} & \textbf{0.74} & \textbf{0.74} & \textbf{0.75} & \textbf{0.76} & \textbf{0.76} & \textbf{0.73} & \textbf{0.75} & \textbf{0.74} & \textbf{0.76} & \textbf{0.75} & \textbf{0.77} & \textbf{0.77} & \textbf{0.76} & 0.72 & \textbf{0.75} & \textbf{0.75} & \textbf{0.74} & \textbf{0.76} & \textbf{0.75} & \textbf{0.75} & \textbf{0.75} & \textbf{0.75} \\
 &  & - & \textbf{0.75} & \textbf{0.77} & \textbf{0.74} & \textbf{0.74} & 0.74 & 0.75 & \textbf{0.76} & \textbf{0.73} & \textbf{0.75} & \textbf{0.74} & \textbf{0.76} & \textbf{0.75} & \textbf{0.77} & \textbf{0.77} & \textbf{0.76} & \textbf{0.75} & \textbf{0.75} & \textbf{0.75} & \textbf{0.74} & 0.75 & 0.74 & \textbf{0.75} & \textbf{0.75} & \textbf{0.75} \\
 \bottomrule
\end{tabular}
}
\caption{BERTScore evaluation of LLM summarization across 23 languages, comparing non-quantized models with 4-bit and 8-bit quantized variants. The best results for each language are in bold.}
\label{tab:summary-quantization}
\end{table*}

\begin{table*}[]
\resizebox{\textwidth}{!}{%
\begin{tabular}{llcllllllllllllllllllllllll}
\midrule
\multicolumn{1}{c}{\textbf{Model}} & \multicolumn{1}{c}{\textbf{Verstion}} & \textbf{Quant} & \multicolumn{1}{c}{\textbf{bg}} & \multicolumn{1}{c}{\textbf{bn}} & \multicolumn{1}{c}{\textbf{ca}} & \multicolumn{1}{c}{\textbf{cs}} & \multicolumn{1}{c}{\textbf{de}} & \multicolumn{1}{c}{\textbf{el}} & \multicolumn{1}{c}{\textbf{en}} & \multicolumn{1}{c}{\textbf{es}} & \multicolumn{1}{c}{\textbf{fi}} & \multicolumn{1}{c}{\textbf{fr}} & \multicolumn{1}{c}{\textbf{hi}} & \multicolumn{1}{c}{\textbf{hu}} & \multicolumn{1}{c}{\textbf{id}} & \multicolumn{1}{c}{\textbf{ko}} & \multicolumn{1}{c}{\textbf{ms}} & \multicolumn{1}{c}{\textbf{my}} & \multicolumn{1}{c}{\textbf{nl}} & \multicolumn{1}{c}{\textbf{pl}} & \multicolumn{1}{c}{\textbf{pt}} & \multicolumn{1}{c}{\textbf{ro}} & \multicolumn{1}{c}{\textbf{sk}} & \multicolumn{1}{c}{\textbf{sr}} & \multicolumn{1}{c}{\textbf{th}} & \multicolumn{1}{c}{\textbf{All}} \\ \midrule
\multirow{6}{*}{Llama3.2 1B Instruct} & \multirow{3}{*}{Article first} & 4bit & 0.21 & 0.16 & 0.20 & 0.17 & 0.22 & 0.19 & 0.28 & 0.20 & 0.11 & 0.17 & 0.06 & 0.06 & 0.21 & 0.24 & 0.15 & 0.14 & 0.22 & 0.14 & 0.15 & 0.17 & 0.18 & 0.14 & 0.04 & 0.17 \\
 &  & 8bit & \textbf{0.24} & 0.26 & \textbf{0.22} & \textbf{0.23} & \textbf{0.26} & 0.25 & \textbf{0.29} & \textbf{0.24} & \textbf{0.19} & \textbf{0.23} & \textbf{0.26} & \textbf{0.10} & \textbf{0.29} & \textbf{0.27} & \textbf{0.27} & \textbf{0.17} & \textbf{0.26} & \textbf{0.22} & \textbf{0.24} & \textbf{0.25} & \textbf{0.22} & \textbf{0.21} & 0.27 & \textbf{0.24} \\
 &  & - & \textbf{0.24} & \textbf{0.27} & \textbf{0.22} & 0.22 & \textbf{0.26} & \textbf{0.26} & \textbf{0.29} & \textbf{0.24} & \textbf{0.19} & 0.22 & \textbf{0.26} & 0.09 & \textbf{0.29} & \textbf{0.27} & \textbf{0.27} & \textbf{0.17} & \textbf{0.26} & \textbf{0.22} & \textbf{0.24} & \textbf{0.25} & \textbf{0.22} & 0.20 & \textbf{0.28} & 0.23 \\ \cline{2-27}
 & \multirow{3}{*}{Article last} & 4bit & 0.01 & 0.00 & 0.04 & 0.05 & 0.04 & 0.02 & 0.27 & 0.06 & 0.02 & 0.05 & 0.01 & 0.03 & 0.06 & 0.05 & 0.04 & 0.02 & 0.07 & 0.02 & 0.05 & 0.04 & 0.03 & 0.01 & 0.01 & 0.04 \\
 &  & 8bit & 0.03 & 0.02 & 0.08 & 0.08 & 0.11 & 0.03 & 0.26 & 0.11 & 0.03 & 0.12 & 0.03 & 0.04 & 0.14 & 0.16 & 0.09 & 0.05 & 0.09 & 0.07 & 0.09 & 0.05 & 0.06 & 0.04 & 0.04 & 0.08 \\
 &  & - & 0.02 & 0.01 & 0.08 & 0.08 & 0.11 & 0.03 & 0.26 & 0.11 & 0.03 & 0.13 & 0.03 & 0.04 & 0.14 & 0.16 & 0.09 & 0.04 & 0.09 & 0.07 & 0.09 & 0.05 & 0.06 & 0.05 & 0.04 & 0.08 \\ \midrule
\multirow{6}{*}{Llama3.2 3B Instruct} & \multirow{3}{*}{Article first} & 4bit & 0.27 & 0.31 & 0.25 & \textbf{0.27} & \textbf{0.28} & 0.27 & \textbf{0.31} & 0.25 & 0.25 & 0.24 & \textbf{0.32} & 0.14 & \textbf{0.32} & 0.30 & \textbf{0.30} & 0.19 & 0.28 & \textbf{0.28} & \textbf{0.22} & \textbf{0.30} & \textbf{0.26} & 0.26 & 0.30 & \textbf{0.27} \\
 &  & 8bit & \textbf{0.29} & 0.31 & 0.26 & \textbf{0.27} & \textbf{0.28} & \textbf{0.29} & \textbf{0.31} & \textbf{0.26} & \textbf{0.28} & \textbf{0.25} & \textbf{0.32} & \textbf{0.17} & \textbf{0.32} & 0.31 & \textbf{0.30} & 0.21 & 0.28 & \textbf{0.28} & 0.20 & \textbf{0.30} & \textbf{0.26} & \textbf{0.27} & \textbf{0.31} & \textbf{0.27} \\
 &  & - & \textbf{0.29} & \textbf{0.32} & \textbf{0.27} & \textbf{0.27} & \textbf{0.28} & \textbf{0.29} & \textbf{0.31} & \textbf{0.26} & \textbf{0.28} & \textbf{0.25} & 0.31 & 0.14 & \textbf{0.32} & \textbf{0.32} & \textbf{0.30} & \textbf{0.22} & \textbf{0.29} & 0.27 & 0.17 & \textbf{0.30} & \textbf{0.26} & 0.26 & \textbf{0.31} & \textbf{0.27} \\ \cline{2-27}
 & \multirow{3}{*}{Article last} & 4bit & 0.26 & 0.30 & 0.19 & 0.22 & 0.24 & 0.24 & 0.29 & 0.23 & 0.18 & 0.24 & 0.29 & 0.06 & 0.30 & 0.30 & 0.28 & 0.19 & 0.25 & 0.24 & 0.12 & 0.21 & 0.22 & 0.21 & 0.29 & 0.23 \\
 &  & 8bit & 0.13 & 0.31 & 0.09 & 0.13 & 0.20 & 0.06 & 0.30 & 0.18 & 0.08 & 0.22 & 0.23 & 0.04 & 0.22 & 0.30 & 0.19 & 0.19 & 0.16 & 0.17 & 0.10 & 0.07 & 0.09 & 0.12 & 0.29 & 0.17 \\
 &  & - & 0.11 & 0.31 & 0.09 & 0.12 & 0.21 & 0.06 & 0.29 & 0.19 & 0.06 & 0.22 & 0.24 & 0.04 & 0.19 & 0.31 & 0.17 & 0.21 & 0.17 & 0.16 & 0.11 & 0.07 & 0.10 & 0.10 & 0.29 & 0.17 \\ \midrule
  \multirow{2}{*}{Llama3.1 70B Instruct} & \multirow{2}{*}{Article first} & 4bit & 0.31 & 0.34 & \textbf{0.29} & 0.29 & \textbf{0.31} & \textbf{0.32} & \textbf{0.33} & \textbf{0.29} & 0.31 & \textbf{0.28} & 0.34 & \textbf{0.31} & \textbf{0.37} & 0.34 & \textbf{0.34} & 0.24 & \textbf{0.32} & \textbf{0.30} & \textbf{0.31} & \textbf{0.33} & 0.28 & \textbf{0.30} & 0.33 & 0.31 \\
 &  & - & \textbf{0.32} & \textbf{0.35} & \textbf{0.29} & \textbf{0.30} & \textbf{0.31} & \textbf{0.32} & \textbf{0.33} & \textbf{0.29} & \textbf{0.32} & \textbf{0.28} & \textbf{0.36} & \textbf{0.31} & \textbf{0.37} & \textbf{0.35} & \textbf{0.34} & \textbf{0.30} & \textbf{0.32} & \textbf{0.30} & \textbf{0.31} & \textbf{0.33} & \textbf{0.29} & \textbf{0.30} & \textbf{0.34} & \textbf{0.32} \\
 \bottomrule
\end{tabular}
}
\caption{ROUGE-L evaluation of LLM summarization across 23 languages, comparing non-quantized models with 4-bit and 8-bit quantized variants. The best results for each language are in bold.}
\label{tab:rouge-quant}
\end{table*}

\section{Veracity Explanations}

\begin{figure*}
    \centering
    \includegraphics[width=\linewidth]{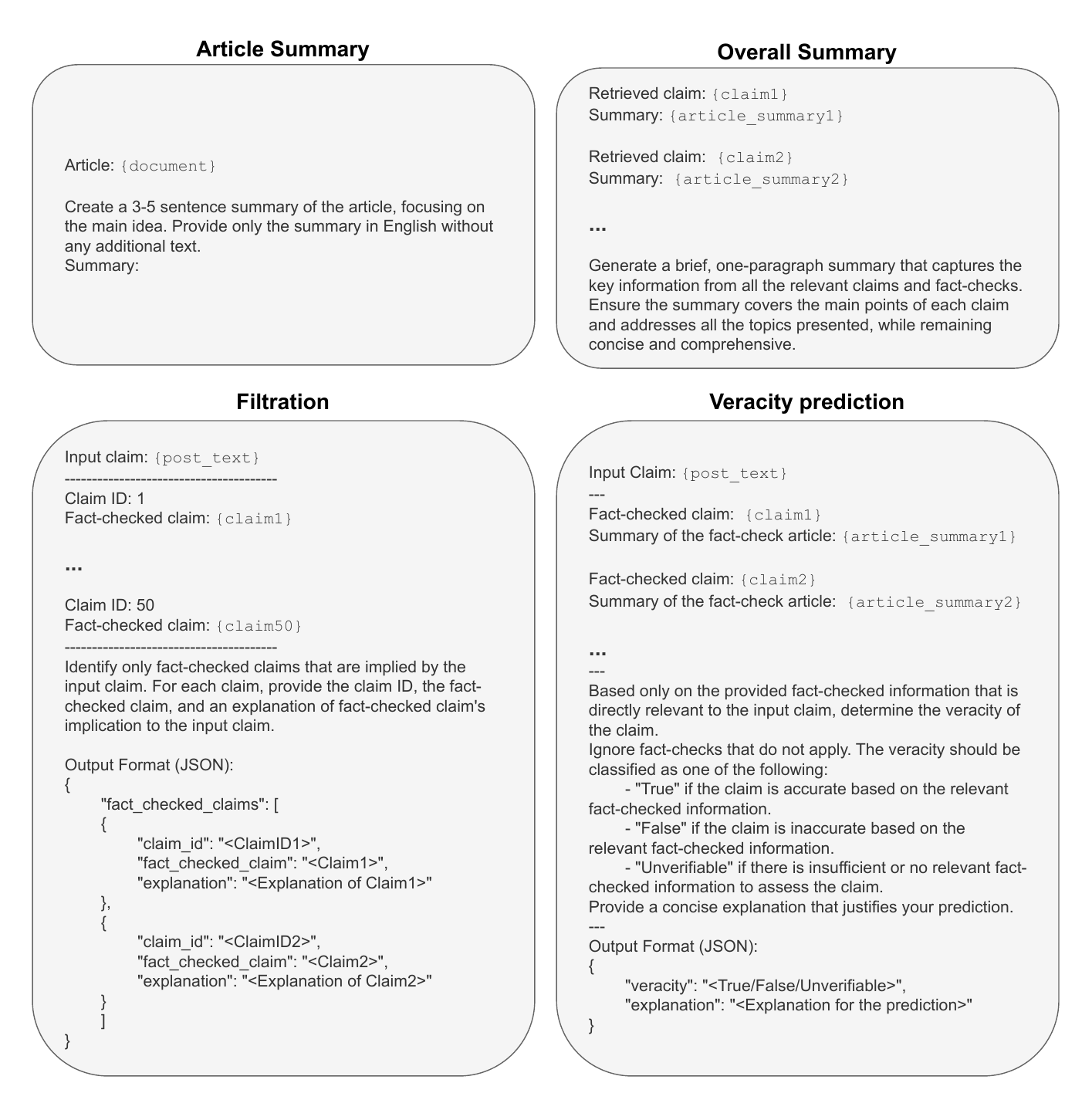}
    \caption{Prompt templates used in the pipeline for the veracity prediction.}
    \label{fig:pipeline_prompts}
\end{figure*}

Figure~\ref{fig:pipeline_prompts} provides the prompt templates for each step within our pipeline. These prompts are used in the pipeline to obtain the final veracity prediction.

\subsection{Error Analysis}

In this section, we investigate the errors and incorrect explanations in veracity prediction. We conducted both manual inspection of a subset of incorrect predictions and automatic analysis to evaluate these errors.

\paragraph{Manual Analysis.} 

For our manual investigation, we randomly selected 20 samples per LLM\footnote{The \texttt{Gemma3} model was not included in the error analysis, as it was added into the study during the final stages, after the manual investigation had already been conducted.} with incorrect predicted labels, resulting in a total of 140 samples. One of the authors analyzed the retrieved relevant fact-checks and LLM-generated explanations, categorizing them into several types.

The most prevalent error category was \textbf{missing context} within claims, which occurred in 27\% of our manually checked samples. This missing context made it difficult to identify relevant fact-checks and predict veracity correctly. Notably, in 63\% of these cases, the LLMs provided correct explanations acknowledging the missing context.

The second most common error (16\% of cases) stemmed from failures in the previous steps, where \textbf{relevant fact-checks were not identified}. In these instances, the LLMs correctly explained that none of the retrieved fact-checked information was directly relevant to the given claim, yet still produced incorrect veracity assessments. 

\textbf{Misunderstanding} of claims and provided relevant fact-checks accounted for 17\% of errors. In these cases, LLMs focused on different aspects of the provided fact-checks or failed to grasp the main point of the claim. We observed some instances where LLMs incorrectly relied on information from the social media post itself to explain its veracity, particularly with longer posts.

Another error pattern (12\% of cases) involved \textbf{LLMs predicting veracity based on ratings mentioned in their generated summaries}, while the actual fact-check ratings differed. For example, a summary might characterize a claim as a hoax, while the rating extracted from the fact-check was "unverifiable".

In 4\% of cases, LLMs relied only on \textbf{the rating from the first fact-check}, despite the presence of fact-checks with correct ratings later in the prompt context. This suggests an incorrect assumption that the first fact-check should be used in veracity prediction.

Finally, 15\% of errors could be attributed to \textbf{ground truth issues}, primarily in cases where fact-checks classified claims as having "no evidence" -- which our normalization process converted to "unverifiable". However, in all these cases, the LLMs' explanations of the claims and their veracity were correct and supported by information from the fact-check summaries.

\paragraph{Automatic Analysis.}

Since one of the observed errors stems from the failure in previous steps to identify relevant fact-checks, we conducted an automatic analysis focusing on the proportion of cases where relevant fact-checks were missing from the retrieved context. Table~\ref{tab:automatic-eror-analysis} presents the percentage of posts for each model where none of the ground truth-relevant fact-checks were included in the list of relevant fact-checks. Without access to relevant fact-checks, models can struggle to accurately predict veracity regardless of their reasoning capabilities. The analysis reseals variations across LLMs, with smaller models generally exhibiting higher rates of missing fact-checks. Notably, \texttt{Qwen2.5 7B} showed the highest proportion (48.6\%) of posts without relevant fact-checks, while \texttt{C4AI Command R+} and \texttt{Mistral Large} performed best with approximately 25\% of posts lacking relevant fact-checks. These findings suggest that retrieval quality remains a bottleneck in the fact-checking pipeline, particularly for smaller models.

\begin{table}
\centering
\small
\begin{tabular*}{\columnwidth}{@{\extracolsep{\fill}}lc@{}}
\toprule
\multicolumn{1}{c}{\textbf{Model}} & \multicolumn{1}{c}{\textbf{\begin{tabular}[c]{@{}c@{}}Missing FC {[}\%{]}\end{tabular}}} \\ 
\midrule
\texttt{Mistral Large} & 25.5 \\ 
\texttt{C4AI Command R+} & 25.3 \\ 
\texttt{Qwen2.5 72B} & 39.1 \\ 
\texttt{Llama3.1 70B} & 33.5 \\ 
\texttt{Llama3.3 70B} & 29.2 \\ \midrule
\texttt{Llama3.1 8B} & 29.9 \\ 
\texttt{Qwen2.5 7B} & 48.6 \\ 
\bottomrule
\end{tabular*}
\caption{Percentage of posts for which no ground truth relevant fact-checks were present in the retrieved context for each LLM.}
\label{tab:automatic-eror-analysis}
\end{table}

\section{Developed Application}


The web-based application integrates the pipeline introduced in Section~\ref{sec:methodology}. For retrieval, we use the best-performing TEM model, \texttt{Multilingual E5}. The backend runs \texttt{Llama3.3 70B}, selected for its strong summarization capabilities and effective filtration of irrelevant fact-checks.

Our fact-check database aggregates fact-checked claims from multiple fact-checking organizations in over 80 languages. We store fact-checked claims, metadata (e.g., language, fact-checking article, rating) and calculated \texttt{Multilingual E5} embeddings of fact-checked claims in Milvus\footnote{\url{https://github.com/milvus-io/milvus}} vector database.

Users submit queries, and the system returns a ranked list of relevant fact-checks identified by the LLM, along with their summaries and explanations. Additionally, the system provides an overall summary, a veracity label distribution graph and an explanation of the verdict. This information supports users in making the final decision.

\subsection{Interface}

\begin{figure*}[t]
    \centering
    \includegraphics[width=\textwidth]{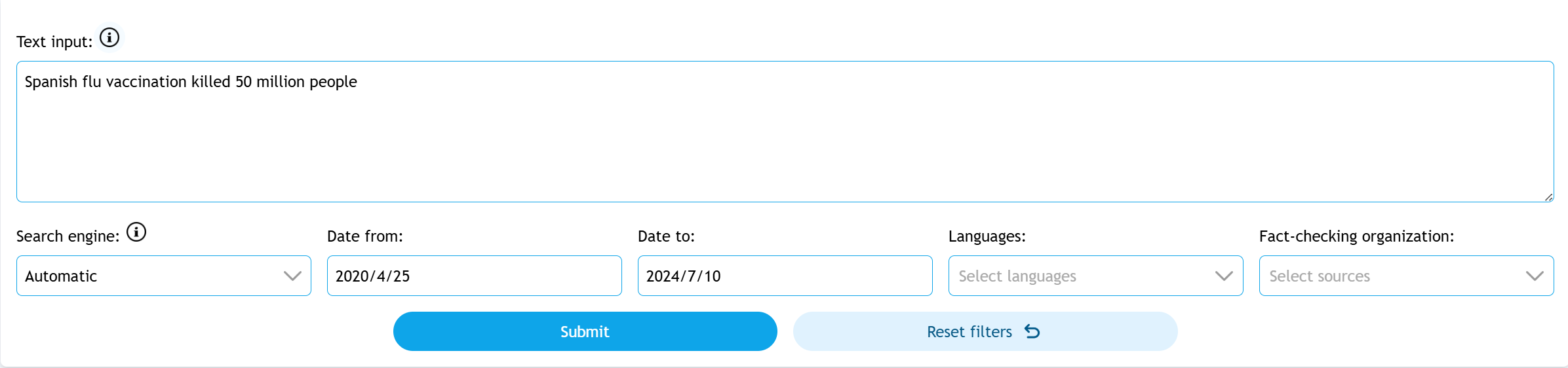}
    \caption{User interface component for the \textbf{text input}.}
    \label{fig:text-input}
\end{figure*}

\begin{figure}[t]
    \centering
    \includegraphics[width=\columnwidth]{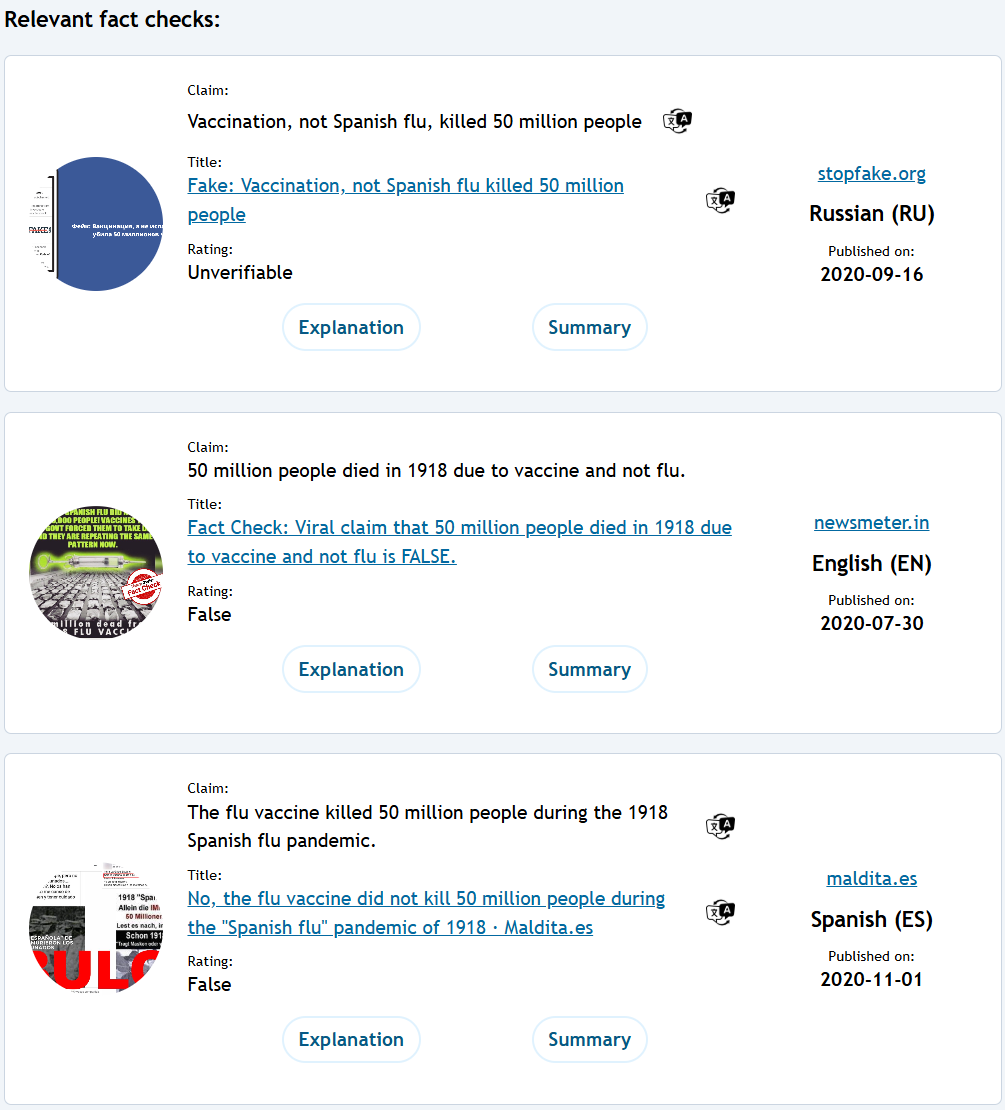}
    \caption{User interface component for a \textbf{list of relevant fact-checks} identified by the LLM within our pipeline. For each relevant fact-check, we provide the summary of the fact-checking article and an explanation of why the fact-check was classified as relevant.}
    \label{fig:relevant_fc}
\end{figure}

\begin{figure}[t]
    \centering
    \includegraphics[width=\columnwidth]{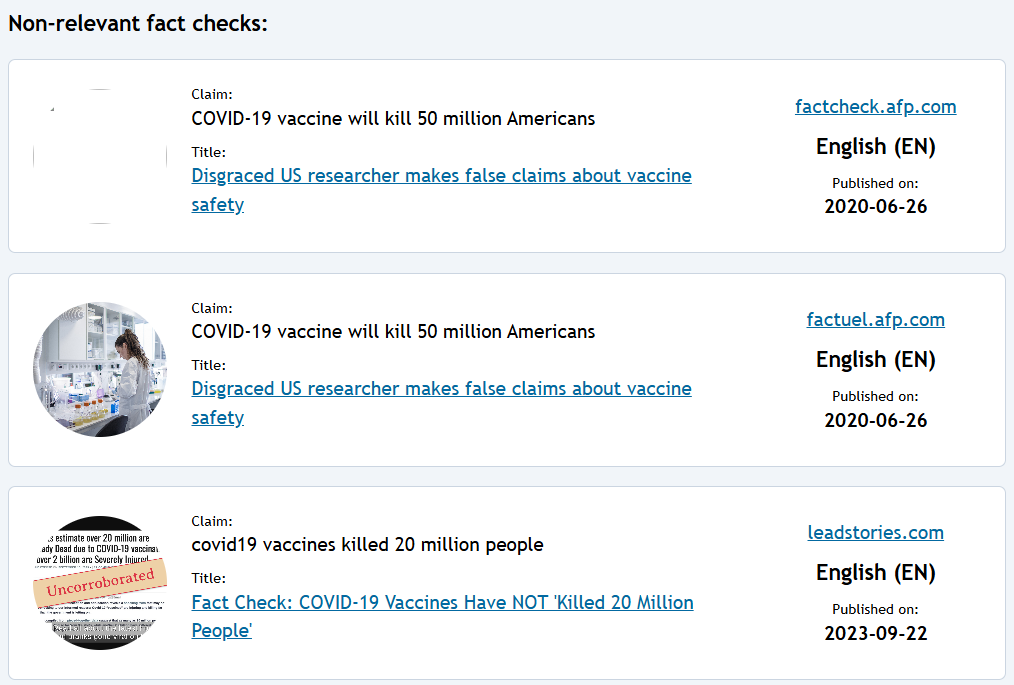}
    \caption{User interface component for a \textbf{list of non-relevant fact-checks}.}
    \label{fig:irrelevant_fc}
\end{figure}

\begin{figure*}[t]
    \centering
    \includegraphics[width=0.75\textwidth]{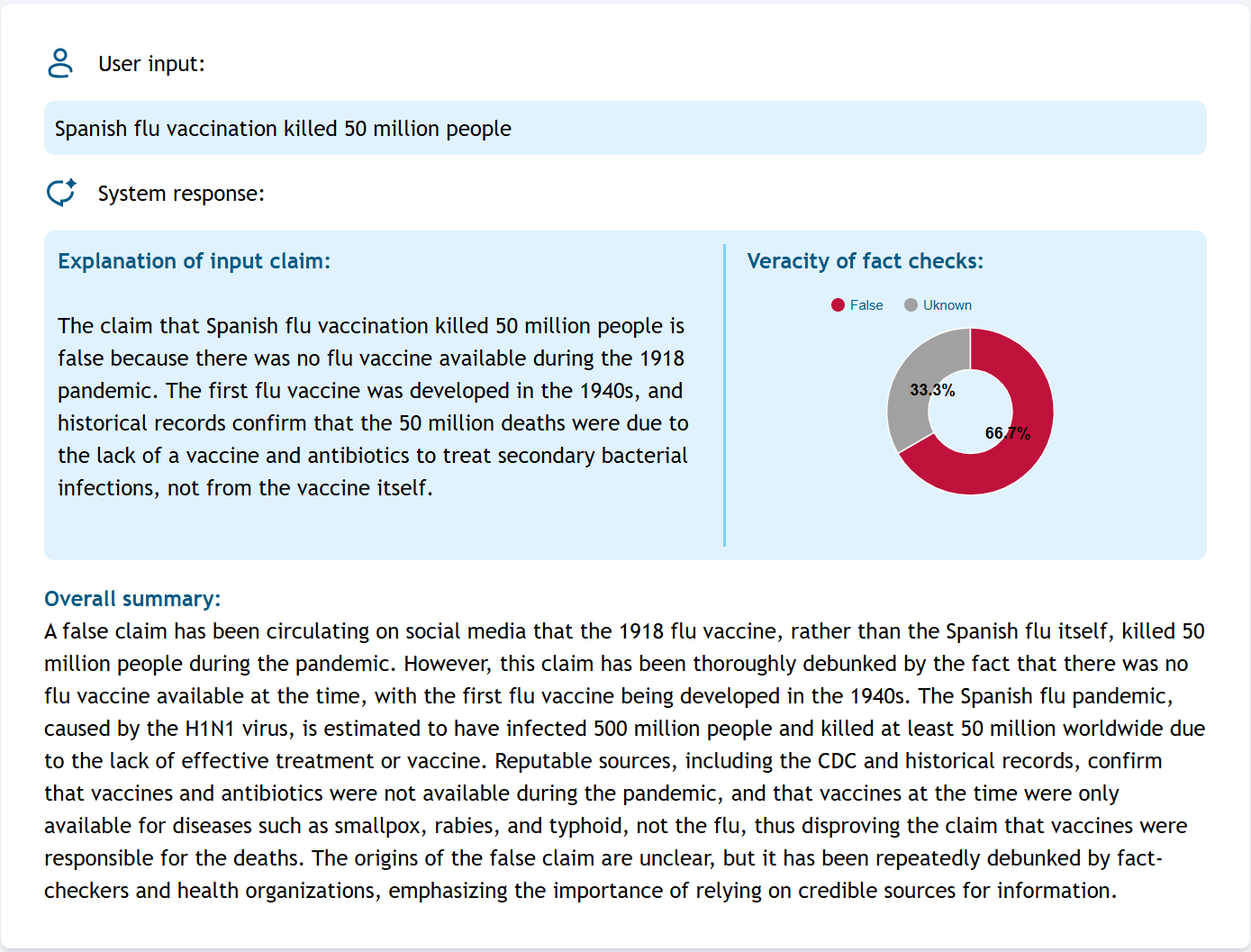}
    \caption{User interface component for \textbf{system response}, where we provide the overall summary of the claim and relevant fact-checks, a veracity distribution graph and the explanation of the predicted veracity prediction.}
    \label{fig:system_response}
\end{figure*}

The developed application consists of four main components. (1) \textit{Text input} (see Figure~\ref{fig:text-input}), where the user provides the claim for which the tool should return relevant fact-checking articles. (2) \textit{List of relevant fact-checks} (see Figure~\ref{fig:relevant_fc}), where we provide all the relevant fact-checks identified by the LLM. (3) \textit{List of non-relevant fact-checks} (see Figure~\ref{fig:irrelevant_fc}), where we list the fact-checks that were retrieved in the retrieval step but were not classified by the LLM as relevant. Since LLMs are not 100\% accurate in identifying relevant fact-checks, we also provide the rest of the fact-checks to make the application robust and provide all the information that was obtained within our pipeline for fact-checkers to make the informed decision. (4) \textit{System response} (see Figure~\ref{fig:system_response}), which includes the overall summary of the input claim and all relevant fact-checks, a veracity distribution graph based on the ratings of the relevant fact-checking articles and an explanation of the predicted veracity label.

\end{document}